\documentclass[runningheads]{llncs}
\usepackage{graphicx}
\usepackage{booktabs}
\usepackage{caption}
\usepackage{subcaption}
\usepackage{amsmath}
\usepackage{multirow}
\usepackage{algorithm}
\usepackage[noend]{algpseudocode}

\usepackage[english]{babel}
\usepackage[utf8]{inputenc}
\usepackage[noend]{algpseudocode}
\usepackage{amsfonts}       

\begin{document}
\title{MMF: A loss extension for feature learning in open set recognition\thanks{Partially supported by grants from Amazon and Rockwell Collins to Philip Chan.}}
%
%
\author{Jingyun Jia\inst{1}\orcidID{0000-0003-0865-049X} \and
Philip K. Chan\inst{1}\orcidID{0000-0002-3878-4205}}
\authorrunning{J. and P.}
%
\institute{Florida Institute of Technology, Melbourne FL 32901, USA}
\maketitle              
\begin{abstract}
The objective of open set recognition (OSR) is to classify the known classes as well as the unknown classes when the collected samples cannot exhaust all the classes. For example, the frequently emerged new malware classes require a system to classify the known classes and identify the unknown malware classes. This paper proposes a loss extension that leverages the neural network to find representations for the known classes so that the representations of the known and the unknown classes become more effectively separable. Our contributions include: First, we introduce an extension that can be incorporated into different loss functions to find more discriminative representations. Second, we show that the proposed extension can significantly improve the performances of two different types of loss functions on datasets from two different domains. Third, we show that with the proposed extension, one loss function outperforms the others in training time and model accuracy.
\end{abstract}
\section{Introduction}
For a multinomial classification problem, the OSR problem's objective is to classify the multiple known classes while identifying the unknown classes. The OSR problem defines a more realistic scenario and has drawn significant attention in application areas such as face recognition \cite{DBLP:journals/cviu/OrtizB14}, malware classification \cite{DBLP:journals/corr/abs-1802-04365} and medical diagnoses \cite{DBLP:conf/ipmi/SchleglSWSL17}. 

In this paper, we introduce an MMF loss extension to help the existing loss functions better handle the open set scenario. The MMF extension is inspired by Extreme Value Signatures (EVS) in \cite{DBLP:conf/dagm/SchultheissKFD17}. Borrowing from a pre-trained neural network for regular classification, EVS uses only the top $K$ activations at one layer for calculating the distance between an instance and a class. The EVS distance function can help identify the unknown class. Instead of using a pre-trained network, we directly learn more discriminative features for the known and unknown classes. We name the approach Min Max Feature (MMF) loss extension because we emphasize the features with the smallest and largest magnitudes during network training. Although the MMF extension is not a standalone loss function, it can be incorporated into different loss functions. Our contribution in this paper is threefold: First, we propose MMF as an extension to different types of loss functions for the OSR problem. Second, we show that MMF achieves statistically significant AUC ROC improvement when applied to two types of loss functions (classification and representation loss functions) on four datasets from two different domains (images and malware). Third, our results indicate that the combination of MMF and the ii loss function \cite{DBLP:journals/corr/abs-1802-04365} outperforms the other combinations in both training time and overall F1 score.

We organize the paper as follows. In section 2, we give an overview of related work. Section 3 presents the MMF loss extension. Section 4 shows that the MMF extension can improve different types of loss functions significantly. 

\section{Related Work}
\label{sec: related_work}
The OSR problem is related to PU (Positive and Unlabeled) learning \cite{DBLP:conf/ecml/LiL05}, which can be regarded as a binary classification problem with the absence of negative samples. The OSR  problem extends the binary classification problem to a multi-class classification problem, with some classes missing from the training set, and will be recognized as an unknown class during testing. We can divide OSR techniques into three categories based on the training set compositions. The first category includes the techniques that borrow additional data in the training set. To better discriminate between known class and unknown class, unlabeled data is introduced during training in \cite{shu2018unseen}. Dhamija et al.\ \cite{DBLP:conf/nips/DhamijaGB18} utilize the differences of feature magnitudes between known and borrowed unknown samples as part of the objective function. Hendrycks et al.\ \cite{DBLP:conf/iclr/HendrycksMD19} propose Outlier Exposure(OE) to distinguish between anomalous (unknown) and in-distribution (known) examples. In general, although borrowing and annotating additional data turns OSR into a common classification problem, the retrieval and selection of additional datasets remain an issue.

The research works that generate additional data in training data fall in the second category of open set recognition techniques. Most data generation methods are based on GANs. Neal et al.\ \cite{neal2018open} add another encoder network to traditional GANs to map from images to a latent space. Lee et al.\ \cite{DBLP:conf/iclr/LeeLLS18} generate ``boundary'' samples in the low-density area of in-distribution acting as unknown samples. While generating unknown samples for the OSR problem has achieved great performance, it requires more complex network architectures. 

The third category of open set recognition does not require additional data. Most of the research works require outlier detection for the unknown class. Pidhorskyi et al.\ \cite{DBLP:conf/nips/PidhorskyiAD18} propose manifold learning based
on training an Adversarial Autoencoder (AAE) to capture the underlying structure of the distributions of known classes. Hassen and Chan \cite{DBLP:journals/corr/abs-1802-04365} propose ii loss for open set recognition. It first finds the representations for the known classes during training and then recognizes an instance as unknown if it does not belong to any known classes. In EVS, Schultheiss et al.\ \cite{DBLP:conf/dagm/SchultheissKFD17} investigate class-specific representations for novelty detection tasks. The research work shows that each class's mean representation can capture discriminative information of both known and unknown classes. EVS focuses on the top $K$ activations via binarizing the activations; however, choosing an appropriate $K$ can be challenging. Also, EVS assumes that all the activation values are positive and only looks at the larger ones. We address both limitations in our proposed approach.

While our proposed approach can be incorporated into different loss functions, we focus on two types of loss functions in this paper: the classification loss functions and the representation loss functions. The objective of classification loss is to lower the classification error of the training data, cross-entropy loss is widely used. The representation loss functions are normally applied to the representation layers, such as triplet loss in \cite{DBLP:conf/cvpr/SchroffKP15} and ii loss in \cite{DBLP:journals/corr/abs-1802-04365}. Triplet loss intends to find an embedding space where the distance between an anchor instance and another instance from the same class is smaller by a user-specified margin than the distance between the anchor instance and another instance from a different class. Ii loss aims to maximize the distance between different classes (inter-class separation) and minimize the distance of an instance from its class mean (intra-class spread). 

\section{Approach}
\label{sec: approach}

We propose the MMF extension to learn more discriminative representations through known classes, thus better separating known and unknown classes. The proposed MMF extension does not borrow or generate additional data for the unknown class, and it can be incorporated into different loss functions. We focus on classification loss functions such as cross-entropy loss and representation loss functions, such as triplet loss and ii loss (Section \ref{sec: related_work}).
 
\begin{figure}[t]

  \begin{minipage}[b]{0.48\textwidth}
     \begin{subfigure}[b]{0.48\textwidth}
\centering
    \includegraphics[width=0.9\linewidth]{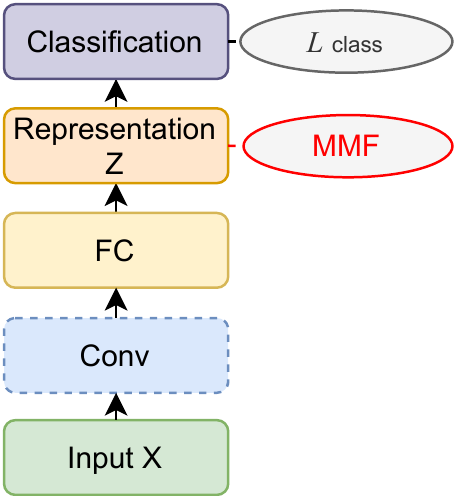}
    \caption{With classification loss}
      \label{fig: mmf-architecture-class}
        \end{subfigure} 
  \begin{subfigure}[b]{0.48\textwidth}           
  \centering
  \includegraphics[width=0.9\linewidth]{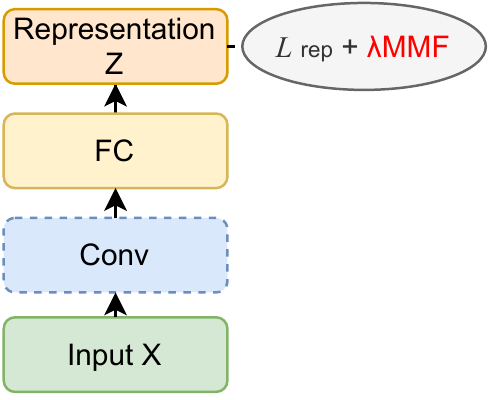}
\caption{With representation loss}
      \label{fig: mmf-architecture-rep}
        \end{subfigure}%
\caption{An overview of the network architectures of different types of loss functions. The convolutional layers are optional. The MMF module in red is our proposed loss extension.}
  \label{fig: mmf-architecture}
      \end{minipage}
        \hfill
    \centering
      \begin{minipage}[b]{0.5\textwidth}

    \includegraphics[width=\linewidth]{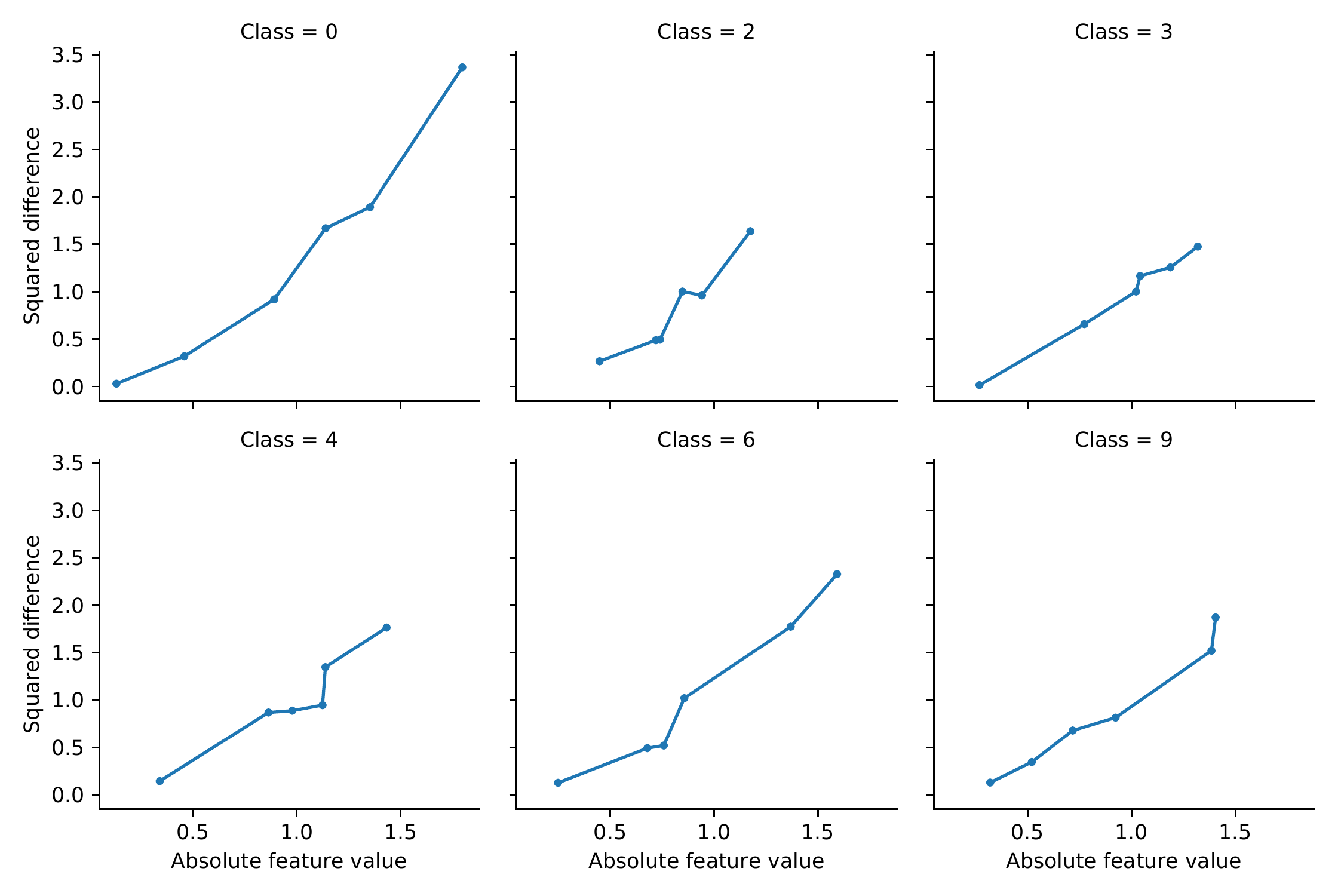}
    \caption{Squared differences of MAV values between the known and unknown classes in Figure \ref{fig: evs-heatmap-heatmap}.}
    \label{fig:ce-diff}
      \end{minipage}

\end{figure}

 \begin{figure*}
\centering
\begin{subfigure}[t]{.33\textwidth}
    \includegraphics[width=\linewidth]{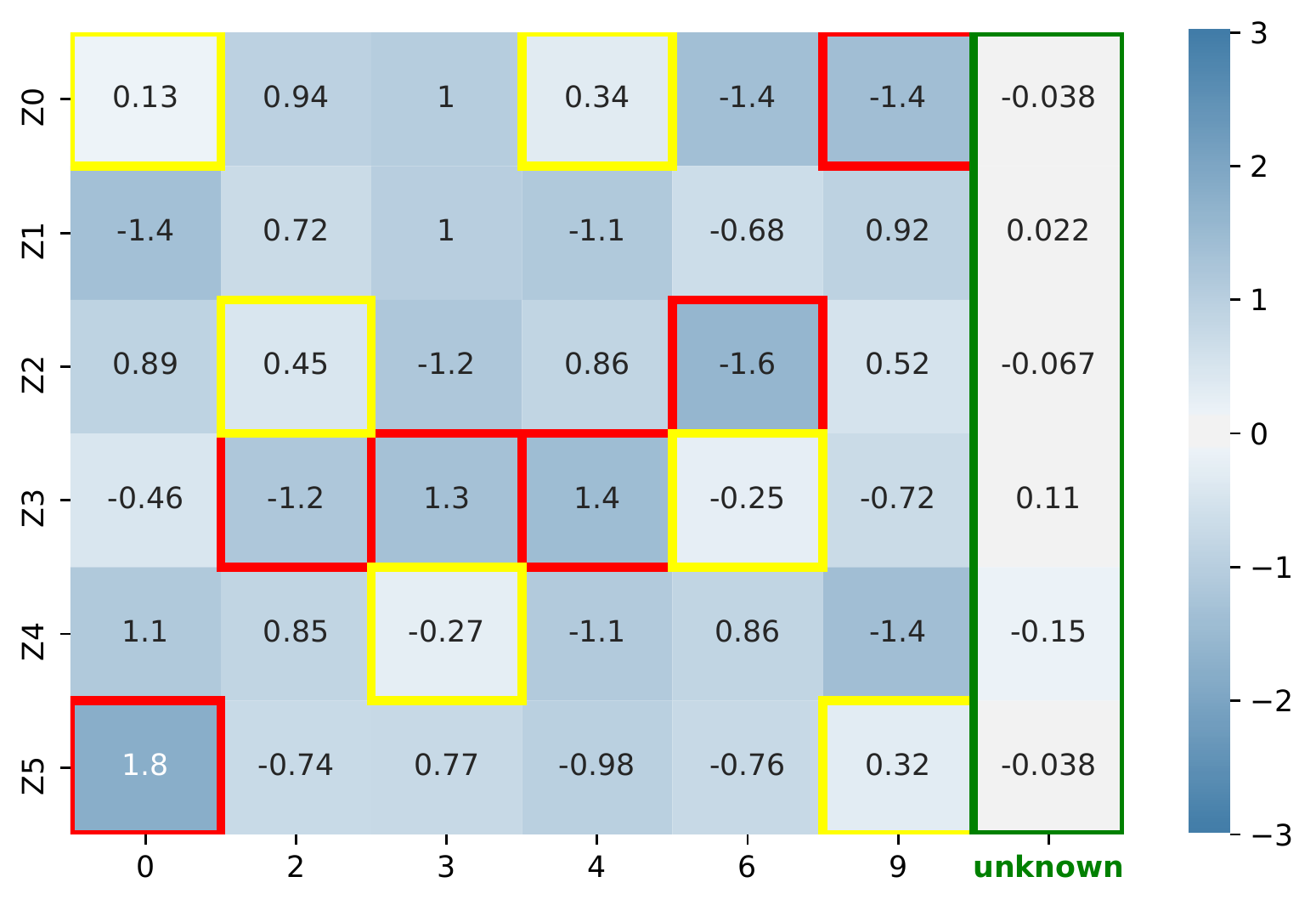}
    \caption{Standalone cross-entropy}
\label{fig: evs-heatmap-heatmap}
\end{subfigure}\hfill
\begin{subfigure}[t]{.33\textwidth}
    \includegraphics[width=\linewidth]{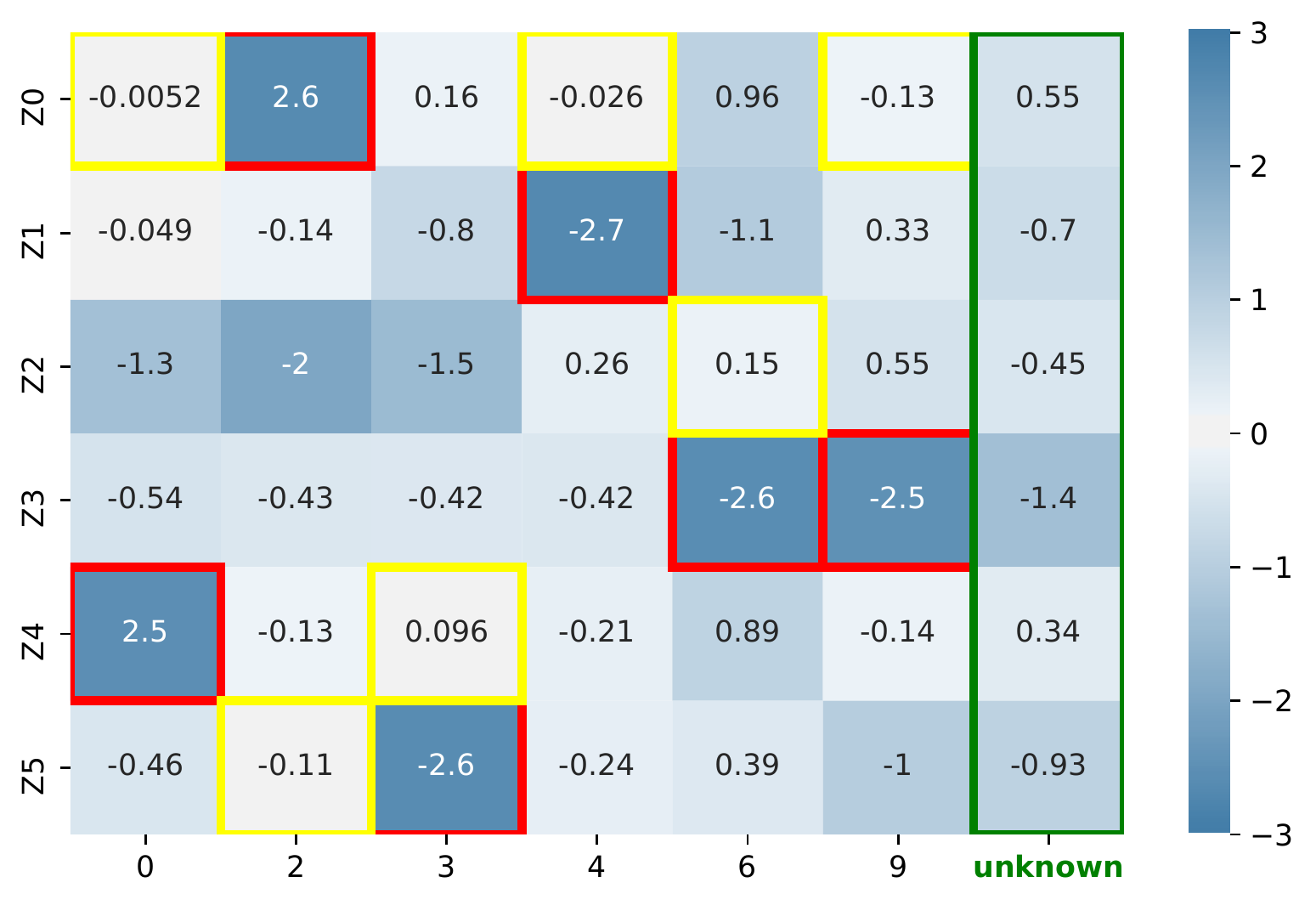}
    \caption{With MaxF}
\label{fig: ce+MaxF-heatmap}
\end{subfigure}\hfill
\begin{subfigure}[t]{.33\textwidth}
    \includegraphics[width=\linewidth]{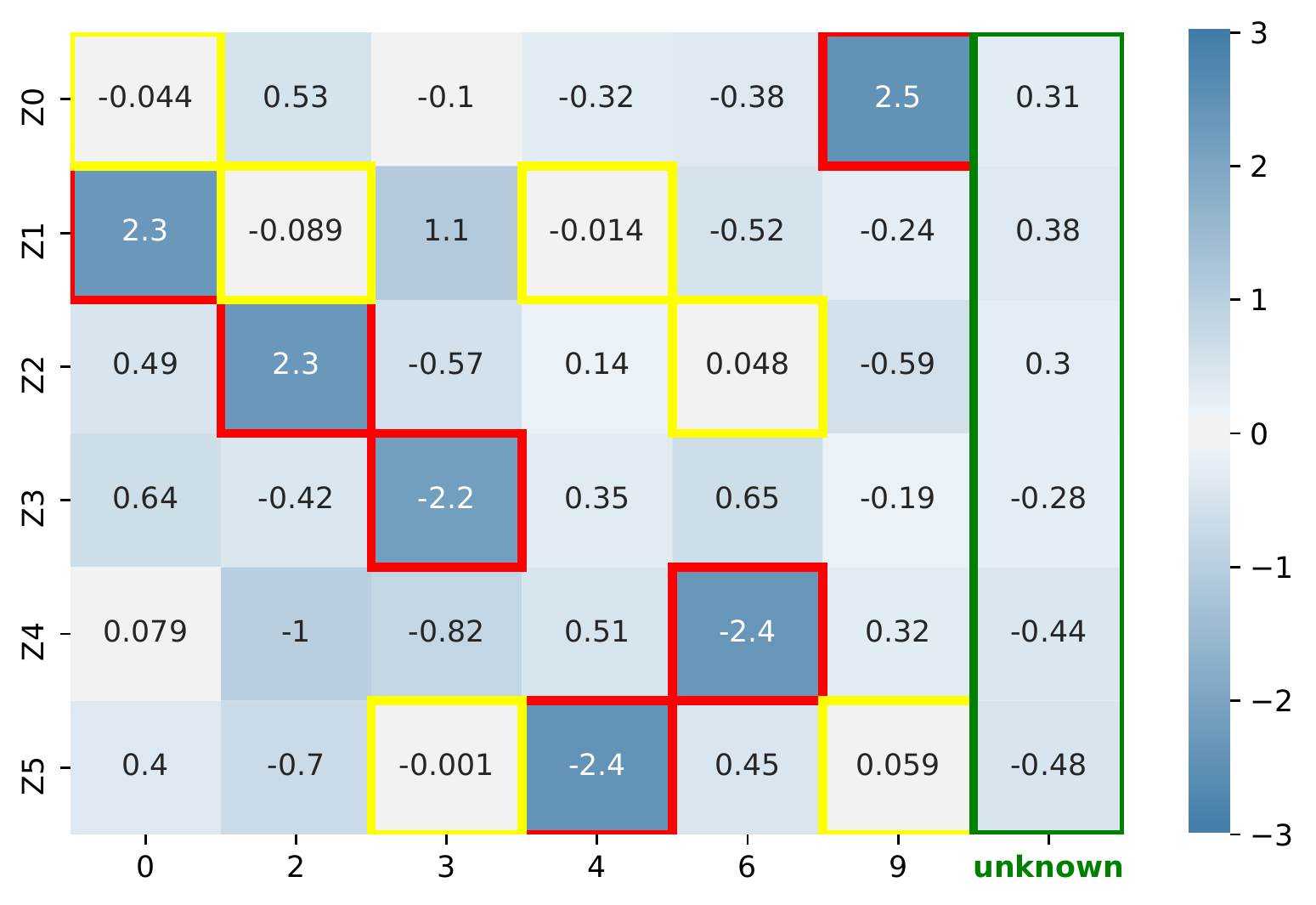}
    \caption{With MMF}
\label{fig: ce+MMF-heatmap} 
\end{subfigure}\hfill
\caption{The heatmap of MAVs (Mean Activation Vectors) of the classes from the MNIST dataset
using cross entropy loss with different extensions.}
\end{figure*}

 A typical classification neural network consists of an input layer, hidden layers, and classification layer. We can consider the hidden layers as different levels of representations of the input. We call the values of the last hidden layer activation vector (AV), and each activation is a learned feature. The mean activation vectors (MAV) of a class is the average of the activation vectors of the class. The network in Figure \ref{fig:  mmf-architecture-class} contains one convolutional layer, one fully connected layer, one representation layer (representation layer Z), and one classification layer (softmax layer).  In some scenarios, a neural network only consists of the input layer and hidden layers as in Figure \ref{fig: mmf-architecture-rep}, where we use learned representations instead of a classification layer for classification tasks. Figure \ref{fig: evs-heatmap-heatmap} shows the learned MAV values from the representation layer using standalone cross-entropy loss. The red boxes are the features with the highest absolute values (magnitudes), the yellow boxes are the features with the lowest absolute values (magnitudes), and the green column is the MAV of the unknown class.
 
 To improve the accuracy of detecting open set samples from unknown classes, we can increase the distances (we use Euclidean distance here) between the learned features of known and unknown samples, summarized by the MAVs of the known and unknown classes. Squared differences are the components of Euclidean distance. Thus we can increase the distance by increasing squared differences. Figure \ref{fig:ce-diff} depicts the relationship between squared differences with the absolute feature values (feature magnitudes) of the six known classes. The x-axis is the absolute feature values in six features, and the y-axis is their corresponding squared differences to the unknown class. We consider a feature with a larger magnitude is more significant than that with a smaller magnitude. We observe that a more significant feature leads to a higher squared difference to the unknown class. The reason is the MAV of the unknown class has a relatively small magnitude as we observe in Figure \ref{fig: evs-heatmap-heatmap}. The small magnitude is due to the unknown class being absent from training and hence its features are not learned. More importantly, the squared difference increases faster with more significant features, which indicates a slight improvement in a more significant feature will increase squared difference more. Thus, we want the features with larger magnitudes to become even more significant to increase the distance between the unknown and known classes. 

However, based on the preliminary experiments, we found that after enlarging the magnitudes of the most significant features for the known classes, the unknown class's MAV became further away from the origin, which reduces the increase in the distance between the known and unknown classes. As shown in Figure \ref{fig: ce+MaxF-heatmap}, the MAV of the unknown class (green column) has significantly increased compared to the one only using standalone cross-entropy loss in Figure \ref{fig: evs-heatmap-heatmap}. To further improve accuracy and increase the magnitudes of the most significant feature, we also decrease the magnitudes of the least significant features to mitigate the increase of the MAV of the unknown class. Comparing Figure \ref{fig: ce+MMF-heatmap} and Figure \ref{fig: evs-heatmap-heatmap}, we can see that after reducing the magnitude of the least significant features, the feature values of unknown classes indeed get smaller.
 
 Therefore, our MMF extension has two properties. Property A maximizes the most significant feature, i.e., the feature with the largest magnitude, for all the known classes. Property B minimizes the least significant feature; i.e., the feature with the smallest magnitude, for all the known classes. As a result, the learned representations for known classes should be more discriminative, while the unknown classes should be less affected. 

\subsection{Learning objectives}
\label{sec:obj}
Let $ \boldsymbol{x} \in \textbf{X}$ be an instance and $y \in {Y}$ be its label. The hidden layers in a neural network can be considered as different levels of representations of input $\boldsymbol{x}$. Suppose that there are $C$ known classes in training data, and $C+1$ classes in test data with the additional class as unknown class. We denote the MAV of class $i$ as $\mu_i$, and $\mu_{ij}$ represents the $j^{th}$ feature of the MAV of class $i$. Assume the AVs and MAVs have $F$ dimensions, representing $F$ features, we stack the MAVs for all the classes to form a representation matrix $\mathbb{U}^{C \times F}$. To satisfy Property A, we first select the most significant features for each class to form the ``max\_feature'' vector. The $i^{th}$ element in ``max\_feature'' is for class $i$:

\begin{equation}
max\_feature_{i} = \max_{1 \leq{j}\leq{F}} | \mu_{ij}|,
\label{eq: max}
\end{equation}

In the example of Figure \ref{fig: evs-heatmap-heatmap}, the ``max\_feature'' would be (1.8, 1.2, 1.3, 1.4, 1.6, 1.4) (the absolute values of the red boxes). Likewise, for Property B, we measure the vector of the ``min\_feature'' as the least significant feature for each class. The $i^{th}$ element is for class $i$:

\begin{equation}
min\_feature_{i} = \min_{1 \leq{j}\leq{F}} | \mu_{ij} |
\label{eq: min}
\end{equation}

The ``min\_feature'' in the example of Figure \ref{fig: evs-heatmap-heatmap} would be (0.13, 0.45, 0.27, 0.34, 0.25, 0.32)  (the absolute values of the yellow boxes). Then, we maximize the smallest value in ``max\_feature'' directly. In this way, all the values in the ``max\_feature'' would be maximized, thus the most significant features for all the known classes would be maximized as Property A. Meanwhile, we minimize the largest value in the ``min\_feature'' to implicitly minimize all the values in the ``min\_feature'', therefore the least significant features for all the known classes would be minimized as Property B. As a result, the proposed MMF extension satisfies both properties:

\begin{equation}
MMF = \max_{1 \leq{i} \leq{C}} (min\_feature_{i}) - \min_{1 \leq{i} \leq{C}}(max\_feature_{i}) \\
\label{eq: minmax}
\end{equation}

In the example of Figure \ref{fig: evs-heatmap-heatmap}, we would like to maximize the ``1.2'' in the ``max\_feature'' and minimize the ``0.45'' in the ``min\_feature''. 
There are alternative methods to generate the ``max\_feature'' and ``min\_feature'', for example, instead of selecting the highest absolute values for ``max\_feature'', we experimented with the highest values ($max\_feature_{1i} = \max_{1 \leq{j}\leq{F}} (\mu_{ij})$) and the lowest values ($max\_feature_{2i} = \max_{1 \leq{j}\leq{F}} (- \mu_{ij})$) to form two ``max\_feature'' vectors and later to be maximized at the same time. However, our experiments indicate that using the single ``max\_feauture'' vector can achieve better performances. There are also other methods to implicitly maximize the most significant features and minimize the least significant values for all the classes, such as maximizing the average value of the ``max\_feature'', or minimizing the average value of the ``min\_feature'', i.e. $\sum_{i=1}^C \frac{1}{C} (min\_feature_i - max\_feature_i)$. However, the results of using average value are weaker than using the extreme values across all classes, hence we choose to use the extreme values in our extension function and in our experiments. 


\subsection{Training with MMF and Open Set Recognition}
\label{sec: loss-function}

In addition to Properties A and B, the MMF extension can be incorporated into different loss functions. We focus on two types of loss functions: a) loss functions designed for decision layers such as cross-entropy loss; b) loss functions designed for representation layers such as triplet loss and ii loss. Notably, we combine the MMF extension with these two types of loss functions differently, as Figure \ref{fig: mmf-architecture}.

We use the network architecture in Figure \ref{fig: mmf-architecture-class} to simultaneously train the network with classification loss functions and the MMF extension. During each iteration, first, we extract AVs and generate the representation matrix; second, we construct the MMF extension function from the ``max\_feature'' vector and ``min\_feature'' vector; third, the weights of each layer of the network are first updated to minimize the MMF extension then updated to minimize classification loss functions using stochastic gradient descent.

 
 The MMF extension can also be incorporated into representation loss functions such as triplet loss and ii loss. As both representation loss functions and the MMF extension should be applied to the layer learning representations, their combination gives us:

\begin{equation}
\mathcal{L} = \mathcal{L}_{rep} + \lambda MMF,
\label{eq: hyper}
\end{equation}
$\mathcal{L}_{rep}$ is a representation loss function, and $\lambda$ is a hyperparameter that strikes a balance between the representation loss function and the MMF extension. 
Figure \ref{fig: mmf-architecture-rep} shows the network architecture using a representation loss function with an MMF extension. The combination serves on the Z-layer of the network. Moreover, the network weights get updated using stochastic gradient descent during each iteration.

After the training process, we obtain the representation centroids for each class. Then during the inference, we use the same strategy as used in ii loss \cite{DBLP:journals/corr/abs-1802-04365}. First, we calculate the outlier score as the distance of learned representation to the nearest representation centroid. Then we sort the outlier score of the training data in descending order and pick the 99 percentile outlier score value as the outlier threshold. If the outlier score of a test sample is beyond the threshold, it will be recognized as the unknown class, otherwise, it will be classified as the known class with the nearest representation centroid.
\section{Experimental Evaluation}

We evaluate the MMF extension with simulated open-set datasets from the following four datasets. 

\noindent\textbf{MNIST}  \cite{DBLP:journals/corr/abs-1802-10135} contains 70,000 handwritten digits from 0 to 9,  which is 10 classes in total. To simulate an open-set dataset, we randomly pick six digits as the known classes participant in the training, while the rest are treated as the unknown class only existing in the test set.

\noindent\textbf{CIFAR-10}  \cite{krizhevsky2009learning} contains 60,000 32x32 color images in 10 classes, with 6,000 images per class. There are 50,000 training images and 10,000 test images. We first convert the color images to grayscale and randomly pick six classes out of the ten classes as the known classes, while instances from the remaining classes are treated as the known class only existing in the test set. 

\noindent \textbf{Microsoft Challenge (MC)} \cite{lecun_cortes_burges} contains disassembled malware samples from 9 families. We use 10260 samples that can be correctly parsed then extract their function call graphs (FCG) as in \cite{DBLP:conf/codaspy/HassenC17} for the experiment. The dimensionality of the FCG is 63x63. Again, to simulate an open-set dataset, we randomly pick six classes as the known classes, while the rest are considered unknowns.

\noindent \textbf{Android Genome (AG)} \cite{zhou_jiang} consists of malicious android apps from many families in different sizes. We use nine families (986 samples) with a relatively larger size for the experiment to be fairly split into the training set, the test set, and the validation set. we first use \cite{DBLP:conf/ccs/GasconYAR13} to extract the function instructions and then extract 1453 raw FCG features as in \cite{DBLP:conf/codaspy/HassenC17}. Like the MNIST and the MC dataset, we randomly pick six classes as the known classes in the training set and consider the rest as the unknown class, which are only used in the test phase.

\subsection{Network Architectures and Evaluation Criteria}
We evaluate the MMF extension associated with two types of loss functions: classification loss functions and representation loss functions. Specifically, we use the cross-entropy loss as the example of classification loss functions, and use ii loss \cite{DBLP:journals/corr/abs-1802-04365} and triplet loss \cite{DBLP:conf/cvpr/SchroffKP15} as the examples of representation loss functions. Moreover, we compare compare these pairs with OpenMax \cite{bendale2016towards}. 

For the MNIST dataset, the padded input layer is of size (32, 32), followed by two non-linear convolutional layers with 32 and 64 nodes. We also use the max-polling layers with kernel size (3, 3) and strides (2, 2) after each convolutional layer. We use two fully connected non-linear layers with 256 and 128 hidden units after the convolutional component. Furthermore, the linear layer Z, where we extract the representation matrix, is six dimensions in our experiment. We use the Relu activation function for all the non-linear layers and set the Dropout's keep probability as 0.2 for the fully connected layers. We use Adam optimizer with a learning rate of 0.001. The network architecture of the CIFAR-10 experiment is similar to the MNIST dataset, except the padded input layer is of size (36, 36). The experiment for the MS Challenge dataset also implements two convolutional layers. The padded input layer is of size (67, 67). However, we only use one fully connected layer instead of two after the convolutional layers. Also, we make the keep probability of Dropout as 0.9. The Android Genome dataset does not use the convolutional component. We use a network with one fully connected layer of 64 units before the linear layer Z. We also used Dropout with a keep probability of 0.9 for the fully connected layers. We set the learning rate of Adam optimizer as 0.1. Besides, we use batch normalization in all the layers to prevent features from getting excessively large. And as mentioned in section \ref{sec: loss-function}, we use contamination ratio of 0.01 for the threshold selection.

 As we discussed in Equation \ref{eq: hyper}, we use a hyperparameter $\lambda$ combine the MMF extension with the representation loss functions (i.e. ii loss and triplet loss in the experiments) as: $\mathcal{L} = \mathcal{L}_{rep} + \lambda MMF$. While the range of $\lambda$ is (0, 1], we set $\lambda$ as 0.2 and 0.5 for ii loss and triplet loss for the MNIST and CIFAR-10 datasets. For the MC dataset, we set $\lambda$ as 0.5 and 0.3 for ii loss and triplet loss. We set $\lambda$ as 0.4 for both ii loss and triplet loss in the AG dataset's experiments. 

For each dataset, we simulate three different groups of open sets then repeat each group 10 runs, so each dataset has 30 runs in total. When measuring the model performance, we use the average AUC scores under 10\% and 100\% FPR (False Positive Rate) for recognizing the unknown class, as lower FPR is desirable in the real world for cases like malware detection. We measure the F1 scores for known and unknown classes ($C+1$ classes) separately as one of the OSR tasks is to classify the known classes. Moreover, we perform t-tests with 95\% confidence in both the AUC scores and F1 scores to see if the proposed MMF extension can significantly improve different loss functions.

\subsection{Experimental Results}


\begin{table*}[t]
\caption{The average AUC scores of 30 runs at 100\% and 10\% FPR of OpenMax and three loss functions quadruples. The underlined values are statistical significant better than the standalone loss functions via t-test with 95\% confidence. The values in bold are the highest values in each quadruple. The values in brackets are the highest values in each row.}
\resizebox{\textwidth}{!}{%
\begin{tabular}{l l  c  cccc cccc cccc}
\toprule
 & \multicolumn{1}{c}{}  & \multicolumn{1}{c}{OpenMax} &\multicolumn{4}{c}{ce} & \multicolumn{4}{c}{ii} & \multicolumn{4}{c}{triplet} \\ \midrule 
  & FPR &  & Standalone & +MMF & +MaxF & +MinF & Standalone & +MMF & +MaxF & +MinF & Standalone & +MMF & +MaxF & +MinF \\  \cmidrule(l){4-7} \cmidrule(l){8-11} \cmidrule(l){12-15} 
\multirow{2}{*}{MNIST} & 100\% & 0.9138 & 0.9255 & \underline{0.9479} & \underline{\textbf{0.9515}} &  0.9393 & 0.9578 & [\underline{\textbf{0.9649}}] & 0.9579 &  0.9607 & 0.9496 & \underline{\textbf{0.9585}}  & 0.9480 & 0.9404 \\
 & 10\% & 0.0590 & \textbf{0.0765} & 0.0744  & 0.0761 & 0.0751 & 0.0821 & [\underline{\textbf{0.0842}}] & 0.0826  & 0.0830 & 0.0750 & \underline{\textbf{0.0796}}& 0.0777  & 0.0739 \\ 
 \multirow{2}{*}{CIFAR-10} & 100\% & [0.6757] & 0.5803 & 0.5982 & \underline{\textbf{0.6103}}  & 0.5807 & 0.6392 & 0.6419 & 0.6437  & \textbf{0.6439} & 0.6106 & \underline{\textbf{0.6248}} & 0.6131  & 0.6127 \\
 & 10\% & 0.0065 & 0.0070 & \underline{0.0089}  & \underline{\textbf{0.0090}} & 0.0077 & [\textbf{0.0103}] & 0.0096 & 0.0100  & 0.0100 & 0.0089 & \underline{\textbf{0.0102}}  & 0.0092 & 0.0093\\ 
\multirow{2}{*}{MC} & 100\% & 0.8739 & 0.9148 & [\underline{\textbf{0.9500}}] & \underline{0.9387}  & \underline{0.9352} & 0.9385 & \underline{ \textbf{0.9461}} & 0.9407  & 0.9397 & 0.9240 & \underline{\textbf{0.9430}} & 0.9317  & 0.9178 \\
 & 10\% & 0.0405 & 0.0530 & \underline{\textbf{0.0635}}  & \underline{0.0600} & \underline{0.0588} & 0.0627 & [\underline{\textbf{0.0656}}] & 0.0629  & 0.0619 & 0.0565 & \underline{\textbf{0.0622}}  & 0.0563 & 0.0546\\ 
\multirow{2}{*}{AG} & 100\% & 0.4150 & 0.7506 & \underline{\textbf{0.8205}}& \underline{0.8152}  & 0.7501 & 0.8427 & 0.8694 & 0.8763 & [\textbf{0.8831}] & 0.8271 & \textbf{0.8379} & 0.8203 & 0.8256 \\
 & 10\% & 0.0010 & 0.0058 & \underline{ 0.0148}  & \underline{\textbf{0.0163}}& \underline{0.0036} & 0.0285 & 0.0305  & [\textbf{0.0368}] & 0.0366& 0.0229 & \textbf{0.0275}  & 0.0260 & 0.0235 \\ \bottomrule

\end{tabular}}
\label{tab:auc}
\end{table*}

We compare the model performances of OpenMax as well as three loss function quadruples: cross-entropy loss, ii loss, and triplet loss. Table \ref{tab:auc} shows the AUC scores of the models in the four datasets; mainly, we focus on comparing the ``Standalone'' with the corresponding ``+MMF'' subcolumns. We observe that the quadruples in general achieve better AUC scores than OpenMax. Moreover, with the MMF extension, the AUC scores of the loss functions have achieved statistically significant improvements in 16 out of 24 cases (3 loss functions$\times$4 datasets$\times$2 FPR values).

\begin{table*}[t]

\caption{The average F1 scores of 30 runs of OpenMax and three loss functions pairs. The underlined values show statistically significant improvements (t-test with 95\% confidence) comparing with the standalone loss functions. The values in bold are the highest values in each column. }
\centering
\resizebox{\textwidth}{!}{%
\begin{tabular}{l l ccc ccc ccc ccc}
\toprule
 & \multicolumn{4}{c}{MNIST} & \multicolumn{3}{c}{CIFAR-10} & \multicolumn{3}{c}{MC} & \multicolumn{3}{c}{AG} \\ \midrule
 & & Known & Unknown & Overall & Known & Unknown & Overall & Known & Unknown & Overall & Known & Unknown & Overall \\ \cmidrule(l){3-5} \cmidrule(l){6-8} \cmidrule(l){9-11}  \cmidrule(l){12-14} 
 \multirow{1}{*}{OpenMax} &  & 0.8964 & \textbf{0.7910} & 0.8814 & \textbf{0.6456} & \textbf{0.5407} & \textbf{0.6306} & 0.8903 & 0.7329 & 0.8679 & 0.2273 & \textbf{0.7761} & 0.3057 \\ \midrule
\multirow{2}{*}{ce}   & Standalone & 0.7596 & 0.7561 & 0.7591 & 0.5672 & 0.3697 & 0.5390 & 0.8881 & 0.6643 & 0.8562 & 0.5346 & 0.0033 & 0.4587 \\
& +MMF & \underline{0.8504} & \underline{0.7902} & \underline{0.8809} & \underline{0.5994} & 0.3271 & \underline{0.5605} & \underline{0.9090} & \textbf{0.7963} & \underline{0.8929} & 0.5555 & \underline{0.1142} & 0.4925 \\
\multirow{2}{*}{ii} & Standalone & 0.9320 & 0.8833 & 0.9250 & 0.6206 & 0.3570 & 0.5829 & 0.9128 & 0.7257 & 0.886 & 0.6349 & 0.6677 & 0.6396 \\
& +MMF & \textbf{0.9373} & 0.8916 & \textbf{0.9308} & 0.6205 & 0.3660 & 0.5842 & \underline{\textbf{0.9210}} & \underline{0.7680} & \underline{\textbf{0.8991}} & \textbf{0.6407} & 0.7251 & \textbf{0.6528} \\ 
\multirow{2}{*}{triplet} & Standalone & 0.9103 & 0.8302 & 0.8989 & 0.5798 & 0.4515 & 0.5614 & 0.8998 & 0.7018 & 0.8715 & 0.5929 & 0.6323 & 0.5986 \\
& +MMF & \underline{0.9239} & \underline{0.8625} & \underline{0.9152} & 0.5943 & 0.4790 & 0.5778 &  0.9064 & 0.7213 & 0.8800 & 0.6005 & 0.6895 & 0.6132 \\ 
 
\bottomrule
\end{tabular}}
\label{tab:f1}
\end{table*}


Table \ref{tab:f1} shows the average F1 scores for the four datasets. We first calculate the F1 scores for each of the $C$ known classes and the unknown class, then average the $C+1$ classes as the Overall F1 scores. We can see the loss functions with the MMF extension have better results than their corresponding standalone versions for both the known and the unknown classes. We observe that ii loss with the MMF extension is more accurate than the other five methods in six out of twelve F1 scores. Particularly, it achieves the highest Overall F1 scores for three out of four datasets.

Table \ref{tab: time} shows the comparison of the average training time of the 30 runs for the MNIST dataset with 5000 iterations via NVIDIA Tesla K80 GPU on AWS. We find that adding the MMF extension almost doubles the training time of using standalone cross-entropy. While for ii loss and triplet loss, adding the extension increases the training time by around $1\%$. The reason is that the MMF extension needs to create the representation matrix from scratch for the network with ce loss, which needs an extra backpropagation step, both of which take more time. We also observe that with our MMF extension, ii loss has the fastest training time among three loss functions. Overall F1 scores and training time indicate that ``ii+MMF'' is the most accurate and efficient combination.



\subsection{Analysis}


Figure \ref{fig: ce+MMF-heatmap} shows the heatmap of MAV values of the simulated open MNIST dataset trained by cross-entropy loss with the MMF extension. We take digits  ``0'', ``2'', ``3'', ``4'', ``6'', ``9'' as the known classes and the remaining digits as the unknown class. Comparing with the MAV values from the network with standalone cross-entropy loss (Figure \ref{fig: evs-heatmap-heatmap}), we can find that the MAVs of the known classes become more discriminative from each other, and each of the known classes has its representative feature. (e.g. Z1 for class ``0'', Z2 for class ``2''). Whereas the MMF extension has less effect on the unknown class, and its MAV values are relatively evenly distributed.

\begin{figure}[t]
\centering
  \begin{minipage}[t]{0.68\textwidth}

 \begin{subfigure}[t]{0.5\textwidth}
 \centering
            \includegraphics[width=\linewidth]{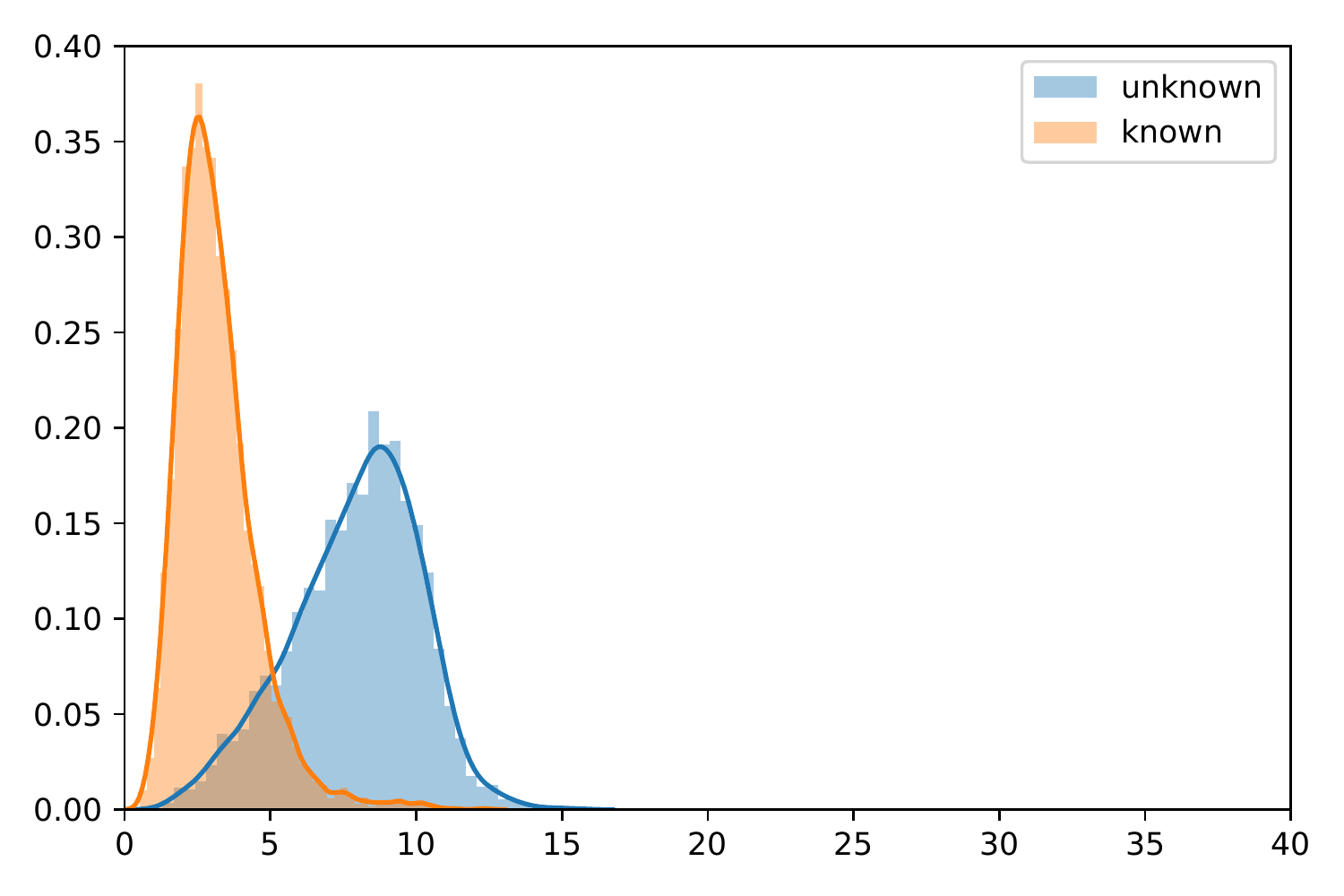}
                \caption{triplet}
        \end{subfigure}%
  \begin{subfigure}[t]{0.5\textwidth}
             \includegraphics[width=\linewidth]{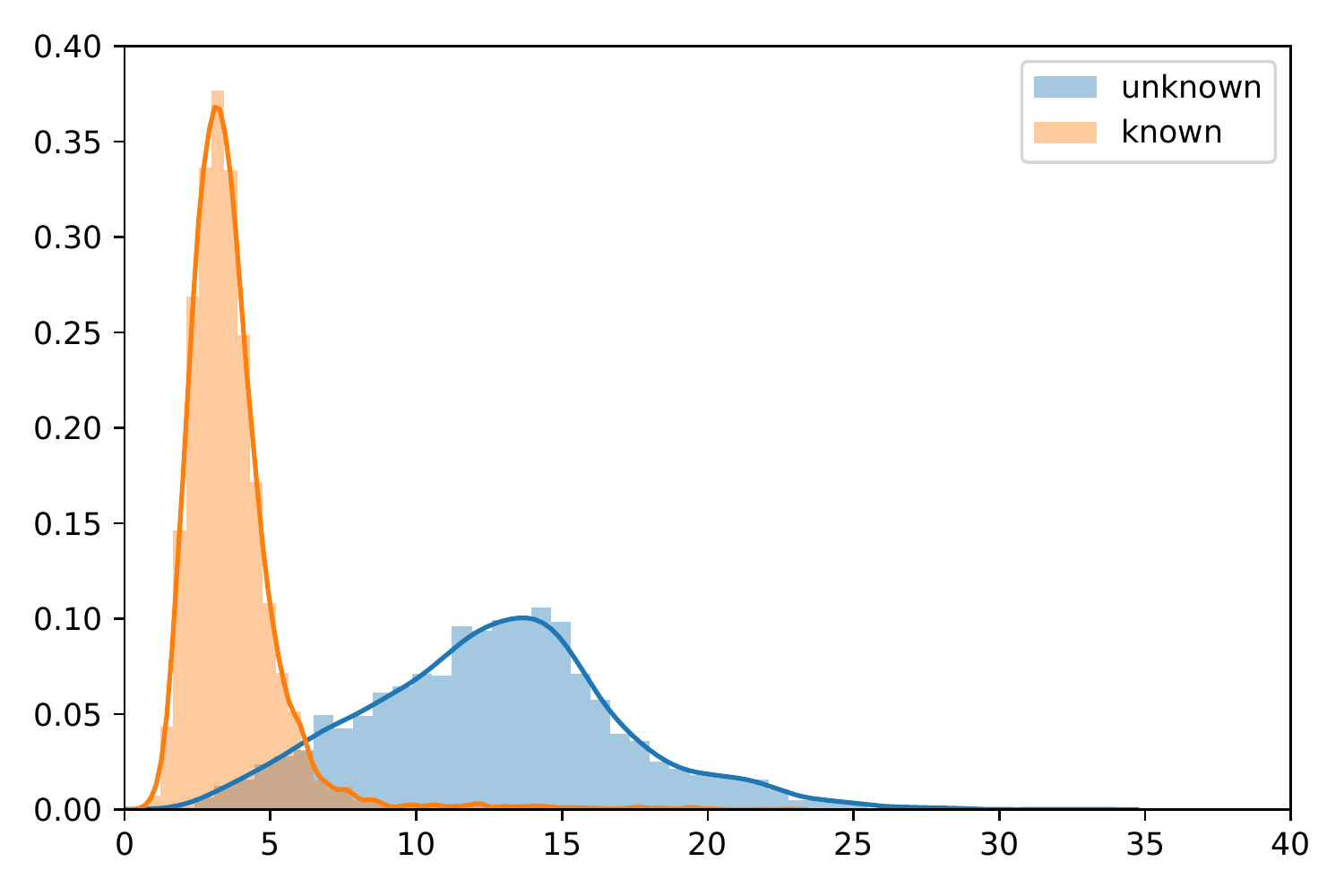}
                \caption{triplet+MMF}
        \end{subfigure}%
\caption{The distributions of outlier scores in MNIST.}
\label{fig: ce-hist}
  \end{minipage}
  \hfill
    \begin{minipage}{0.3\textwidth}
\captionof{table}{The comparison of training time for the MNIST dataset.}
\scalebox{0.75}{
\begin{tabular}[t]{l ccc}
\toprule
 & Regular & +MMF & delta \\ \midrule
ce & 119.33 & 230.43 & +111.1 \\
ii & 122.17 & 123.30 & +1.14 \\
triplet & 223.27 & 225.70 & +2.43 \\ \bottomrule
\label{tab: time}
\end{tabular}}
\end{minipage}
\end{figure}

\begin{figure}[t]
 \begin{subfigure}[b]{0.48\textwidth}
               \includegraphics[width=\linewidth]{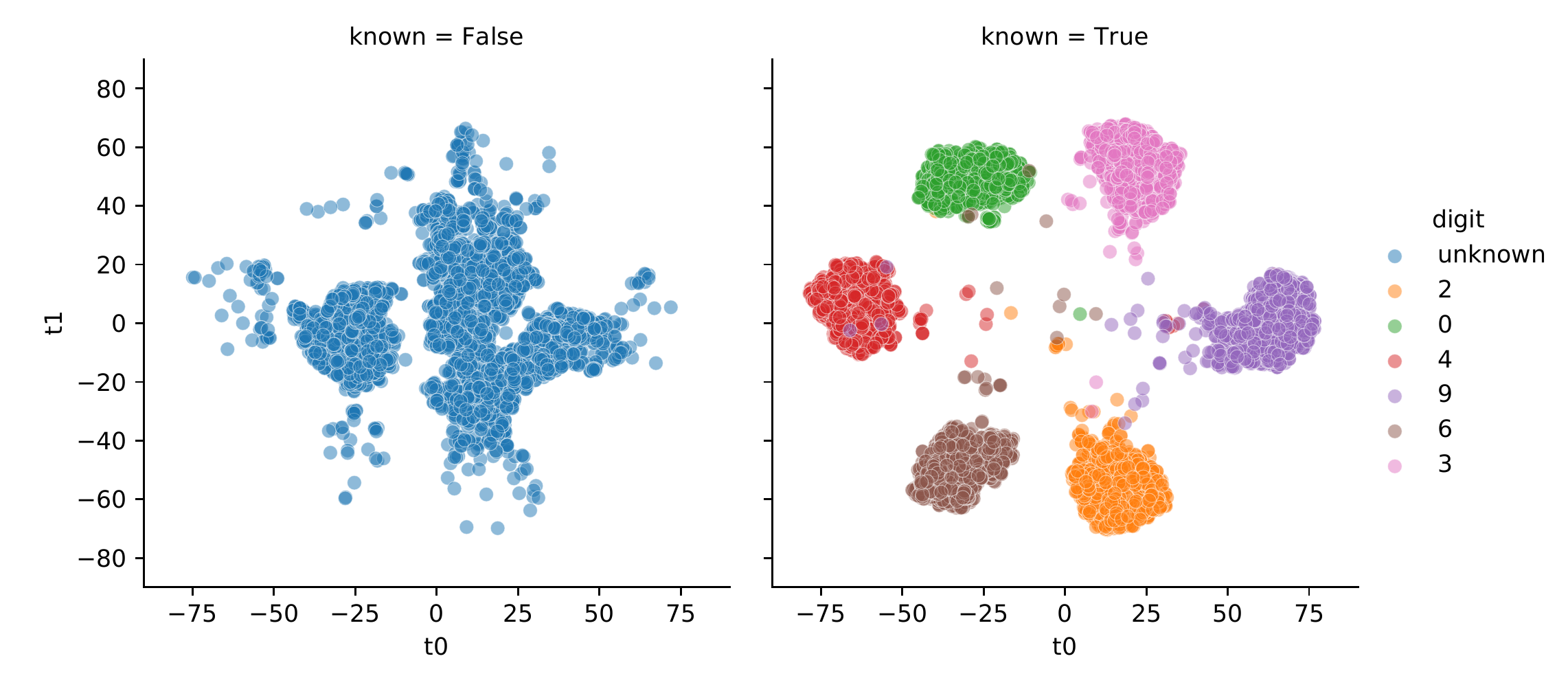}
                \caption{triplet}
        \end{subfigure} 
  \begin{subfigure}[b]{0.48\textwidth}                \includegraphics[width=\linewidth]{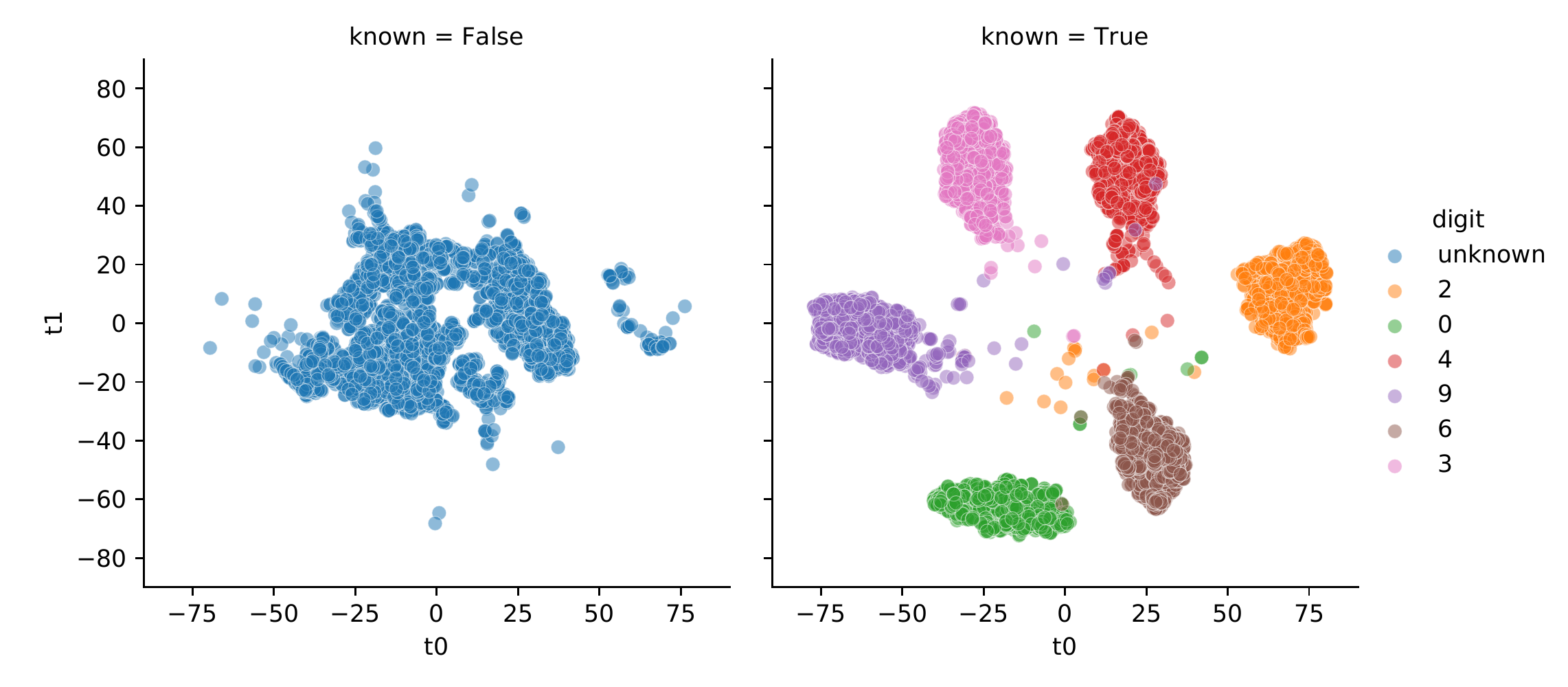}
                \caption{triplet+MMF}
        \end{subfigure}%
\caption{The t-SNE plots of the MNIST dataset in the experiments of triplet vs. triplet+MMF. The left subplots of (a) and (b) are the representations of the unknown class (a mixture of digits ``1'', ``5'', ``7'' and ``8''), and the right plots are the representations of the known classes.}
\label{fig: ce-tsne}
\end{figure}

Since we recognize the unknown class based on the outlier score described in section 3.3, we analyze both the test samples' outlier scores from the known classes and the unknown class from the MNIST experiment. Figure \ref{fig: ce-hist} shows the histogram of the distributions of the outlier scores in triplet loss experiments and triplet loss with the MMF extension. Compared with using standalone triplet loss, adding an MMF extension increases the outlier scores of the unknown class, which pushes the score distributions further away from those of the known classes and results in fewer overlaps between the known classes the unknown class. It is the reduced overlaps that make the known classes and the unknown classes more separable than before. Figure \ref{fig: ce-tsne} shows the t-SNE (perplexity: 50) plots of the Z-layer representations of the MNIST dataset from the same experiments. We can see that with the MMF extension, not only the known classes and the unknown class are more separate from each other, the known classes become more disparate than before.

We also perform an ablation analysis for the MMF loss extension to understand the importance of the MMF extension's two properties. As shown in Table \ref{tab:auc}, our baselines include (1) standalone loss functions; (2) loss functions with an extension that maximize the most significant feature as Property A (MaxF); (3) loss functions with an extension that minimizes the least significant feature as Property B (MinF). In general, the MMF extension with both properties outperforms the baselines. This result is consistent with our motivation for the two properties at the beginning of Section \ref{sec: approach}. Moreover, we find that MaxF and MinF extensions can also achieve better performance than standalone loss functions. While both properties improve AUC scores, Property A (MaxF) has a larger improvement. Hence, Property A plays a more critical role in AUC improvement than Property B.

\begin{figure}[t]
\centering
\begin{subfigure}[t]{.2\textwidth}
    \includegraphics[width=0.7\linewidth]{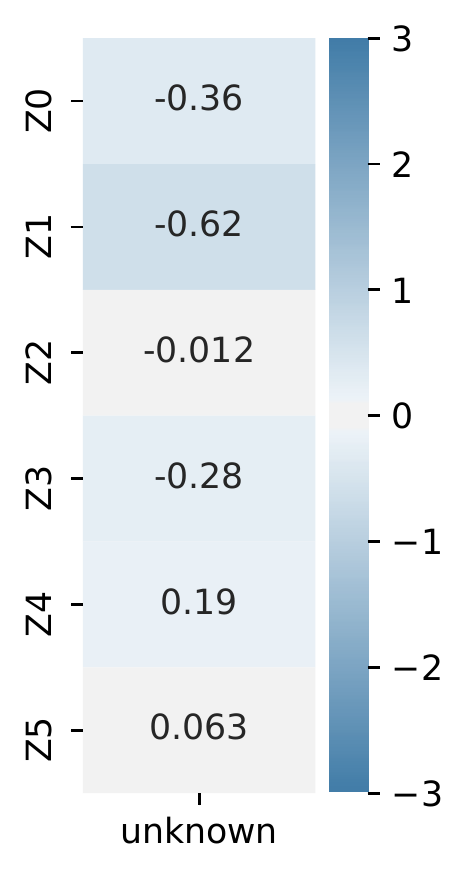}
    \caption{Standalone}
\label{fig: unknown-regular}
\end{subfigure}\hfill
\begin{subfigure}[t]{.2\textwidth}
    \includegraphics[width=0.7\linewidth]{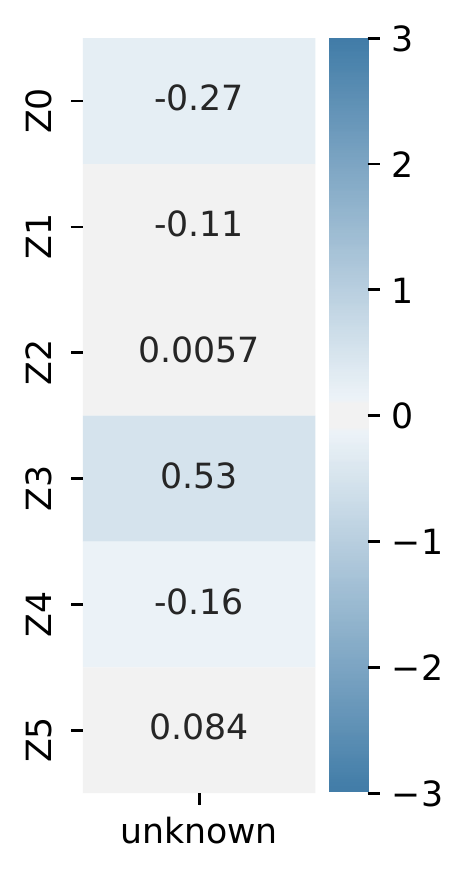}
    \caption{+MinF}
\label{fig: unknown-minf}
\end{subfigure}\hfill
\begin{subfigure}[t]{.2\textwidth}
    \includegraphics[width=0.7\linewidth]{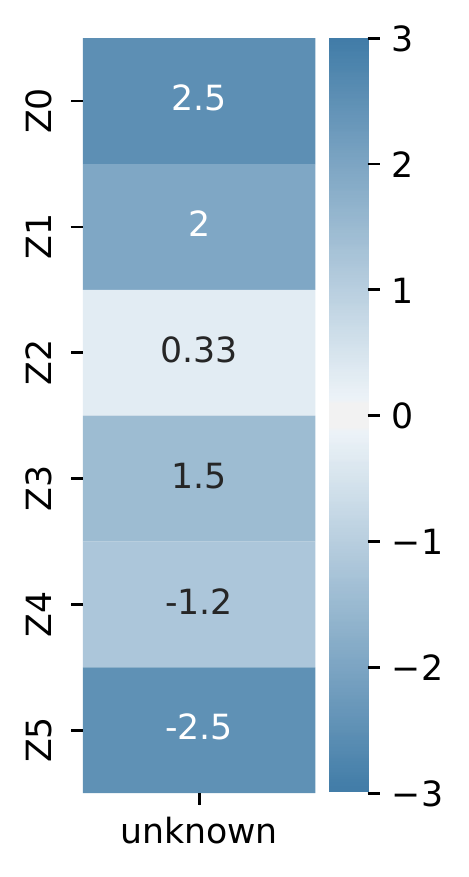}
    \caption{+MaxF}
\label{fig: unknown-maxf} 
\end{subfigure}\hfill
\begin{subfigure}[t]{.2\textwidth}
    \includegraphics[width=0.7\linewidth]{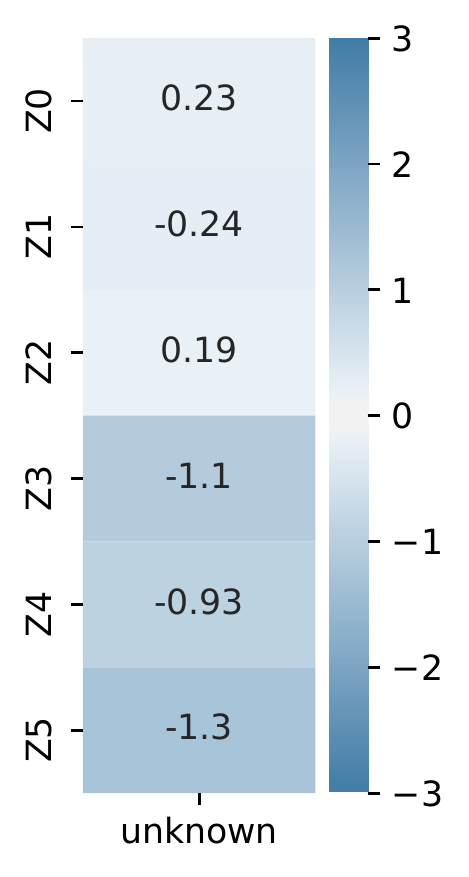}
    \caption{+MMF}
\label{fig: unknown-mmf} 
\end{subfigure}\hfill
\caption{The heatmap of the unknown class's MAV in the experiment of cross entropy loss (ce) on the Microsoft Challenge dataset (MC).}
\label{fig: MC}
\end{figure}

To investigate why MinF also helps improve AUC performance, we show the heatmap of the MAV for the unknown class in the experiment of ce on the MC dataset in Figure \ref{fig: MC}. Comparing Figure \ref{fig: unknown-regular} and Figure \ref{fig: unknown-minf}, we observe that MinF reduced the feature magnitudes for the unknown class, thus increased the distance between the known and unknown classes. Similarly, from Figure \ref{fig: unknown-maxf} and Figure \ref{fig: unknown-mmf}, we observe that the feature magnitudes of the unknown class in MMF (MaxF+MinF) are much smaller than the ones in MaxF. The second observation is consistent with the earlier discussion on adding MinF to help MaxF in MMF at the beginning of Section \ref{sec: approach}. 

\section{Conclusion}
We introduced an add-on loss function extension for the OSR problem. The extension maximizes the feature with the largest magnitude meanwhile minimizes the one with the smallest magnitude for all the known classes during training so that the learned presentations are more discriminative from each other. We have shown that while the known classes are more discriminative from each other, the feature values of unknown classes are less affected by the extension, hence simplifying the open set recognition. We incorporated the proposed extension into both classification and representation loss functions and evaluated them in images and malware samples. The results show that the proposed approach has achieved statistically significant improvements for different loss functions.

\bibliographystyle{splncs04}
\bibliography{main}
\end{document}


\title{\Large Supplementary materials for \\ MMF: A loss extension for feature learning in open set recognition\thanks{Partially supported by grants from Amazon and Rockwell Collins to Dr. Philip Chan.}}
\author{Jingyun Jia \thanks{Florida Institute of Technology. jiaj2018@my.fit.edu}
\and Philip K. Chan \thanks{Florida Institute of Technology. pkc@cs.fit.edu}}

\date{}
\maketitle

The supplementary materials include four parts, and we organize them as follows: In section A, we experiment with an alternative approach to generate ``max\_feature'', namely ``dual max\_feature''. Section B performs more analysis on the MNIST dataset beyond section 4.3 in the paper. Section C provides more details about the datasets and source code that we used for experiments in the paper. Section D discusses potential future research directions. 

\section{Dual max\_feature}

In the paper, we select the highest absolute feature values in the MAVs for each class to form the ``max\_feature'', as $max\_feature_{i} = \max_{1 \leq{j}\leq{F}} | \mu_{ij}|$. In this way, either the highest feature value or the lowest feature value will be extracted. An alternative approach is that we select both the highest and the lowest feature values to form a ``dual max\_feature''. We extract the highest feature value as the first factor of ``dual max\_feature'' as:
\begin{equation*}
   max\_feature\_h_{i} = \max_{1 \leq{j}\leq{F}} (\mu_{ij}) 
\end{equation*}
Meanwhile we extract the lowest feature value as the second factor as:
\begin{equation*}
       max\_feature\_l_{i} = \max_{1 \leq{j}\leq{F}} (- \mu_{ij})
\end{equation*}
In the example of Figure 1c in the paper, the ``dual max\_feature'' consists of two parts:  $max\_feature\_h$ and $max\_feature\_l$. $max\_feature\_h$ would be (1.6, 1.6, 1.2, 0.78, 1.3, 0.73), and $max\_feature_l$ would be (0.78, 0.59, 1.4, 1.4, 1, 1.5). Next, as in the paper, we maximize the smallest values in $max\_feature_h$ and $max\_feature_l$ directly to emphasize both the highest and the lowest feature values for all the known classes:

\begin{equation*}
\begin{split}
 MaxF2 = & - \min_{1 \leq{i} \leq{C}} (max\_feature\_h_{i}) \\ & - \min_{1 \leq{i} \leq{C}}(max\_feature\_l_{i})   
\end{split}
\end{equation*}

Combine the ``dual max\_feature'' with the ``min\_feature'', we have:
\begin{equation*}
\begin{split}
 MMF2 = & \max_{1 \leq{i} \leq{C}} (min\_feature_{i}) \\
 & - \min_{1 \leq{i} \leq{C}}(max\_feature\_h_{i}) \\ & - \min_{1 \leq{i} \leq{C}}(max\_feature\_l_{i})   
\end{split}
\end{equation*}

The results in Table  \ref{tab:dual_max_auc} shows the AUC scores under the ROC curve of the models in the three datasets, with different False Positive Rate (FPR) using the same network architectures in the paper. We can see that using ``dual max\_feature'' does not improve the ROC scores in ``+MaxF'' and the ``+MMF'' columns significantly, and the ``+MMF'' columns still outperform the other methods in 12 out of 24 cases (3 loss functions$\times$ 4 datasets$\times$2 FPR values). Thus we choose to use ``max\_feature'' instead of ``dual max\_feature'' in the paper.

\begin{table*}[h]
\caption{The average AUC scores of 30 runs at 100\% and 10\% FPR of different approaches: +MMF (MMF using ``max\_feature'' in paper), +MMF2 (MMF using ``dual max\_feature''), +MaxF (MaxF using ``max\_feature'' in paper), +MaxF2 (MaxF using ``dual max\_feature''). The values in bold are the
highest value in each quadruple.}
\resizebox{\textwidth}{!}{%
\begin{tabular}{l l cccc cccc cccc}
\toprule
 & \multicolumn{5}{c}{ce} & \multicolumn{4}{c}{ii} & \multicolumn{4}{c}{triplet} \\ \midrule 
  & FPR & +MMF & +MMF2 & +MaxF & +MaxF2 & +MMF & +MMF2 & +MaxF & +MaxF2 & +MMF & +MMF2 & +MaxF & +MaxF2 \\ \cmidrule(l){3-6} \cmidrule(l){7-10} \cmidrule(l){11-14} 
\multirow{2}{*}{MNIST} & 100\% & 0.9479 & 0.9418 & \textbf{0.9515} & 0.9445 & \textbf{0.9649} & 0.9600 & 0.9579 & 0.9595 & \textbf{0.9585} & 0.9528 & 0.9480 & 0.9553 \\
 & 10\% & 0.0744 & 0.0723 & \textbf{0.0761} & 0.0740 & \textbf{0.0842} & 0.0828 & 0.0826 & 0.0829 & \textbf{0.0796} & 0.0789 & 0.0777 & 0.0794 \\
 \multirow{2}{*}{CIFAR-10} & 100\% & 0.5982 & 0.6042 &  0.6103  & \textbf{0.6128} &  0.6419 & 0.6391 & \textbf{0.6437} & 0.6416 & \textbf{0.6248} & 0.6206 & 0.6131 & 0.6189 \\
 & 10\% & 0.0089 & 0.0097 & 0.0090 & \textbf{0.0108} & 0.0096 & 0.0101 & 0.0100 & \textbf{0.0103} & \textbf{0.0102}  & 0.0097 & 0.0092 & 0.0098 \\
\multirow{2}{*}{MC} & 100\% & \textbf{0.9500} & 0.9346 & 0.9387 & 0.9330 & \textbf{0.9461} & 0.9375 & 0.9407 & 0.9388 & \textbf{0.9430} & 0.9304 & 0.9317 & 0.9427 \\
 & 10\% & \textbf{0.0635} & 0.0605 & 0.0600 & 0.0596 & \textbf{0.0656} & 0.0616 & 0.0629 & 0.0616 & \textbf{0.0622} & 0.0588 & 0.0563 & 0.0604 \\
\multirow{2}{*}{AG} & 100\% & 0.8205 & 0.8085 & 0.8152 & \textbf{0.8206} & 0.8694 & 0.8713 & \textbf{0.8763} & 0.8679 & 0.8379 & 0.8368 & 0.8203 & \textbf{0.8390} \\
 & 10\% & 0.0148 & 0.0148 & 0.0163 & \textbf{0.0216} & 0.0305 & 0.0318 & \textbf{0.0368} & 0.0356 & 0.0275 & \textbf{0.0277} & 0.0260 & 0.0267 \\ \bottomrule
\end{tabular}}
\label{tab:dual_max_auc}
\end{table*}
\section{Further Analysis on the MNIST dataset}

\begin{figure}[ht]
 \begin{subfigure}[b]{0.25\textwidth}
               \includegraphics[width=\linewidth]{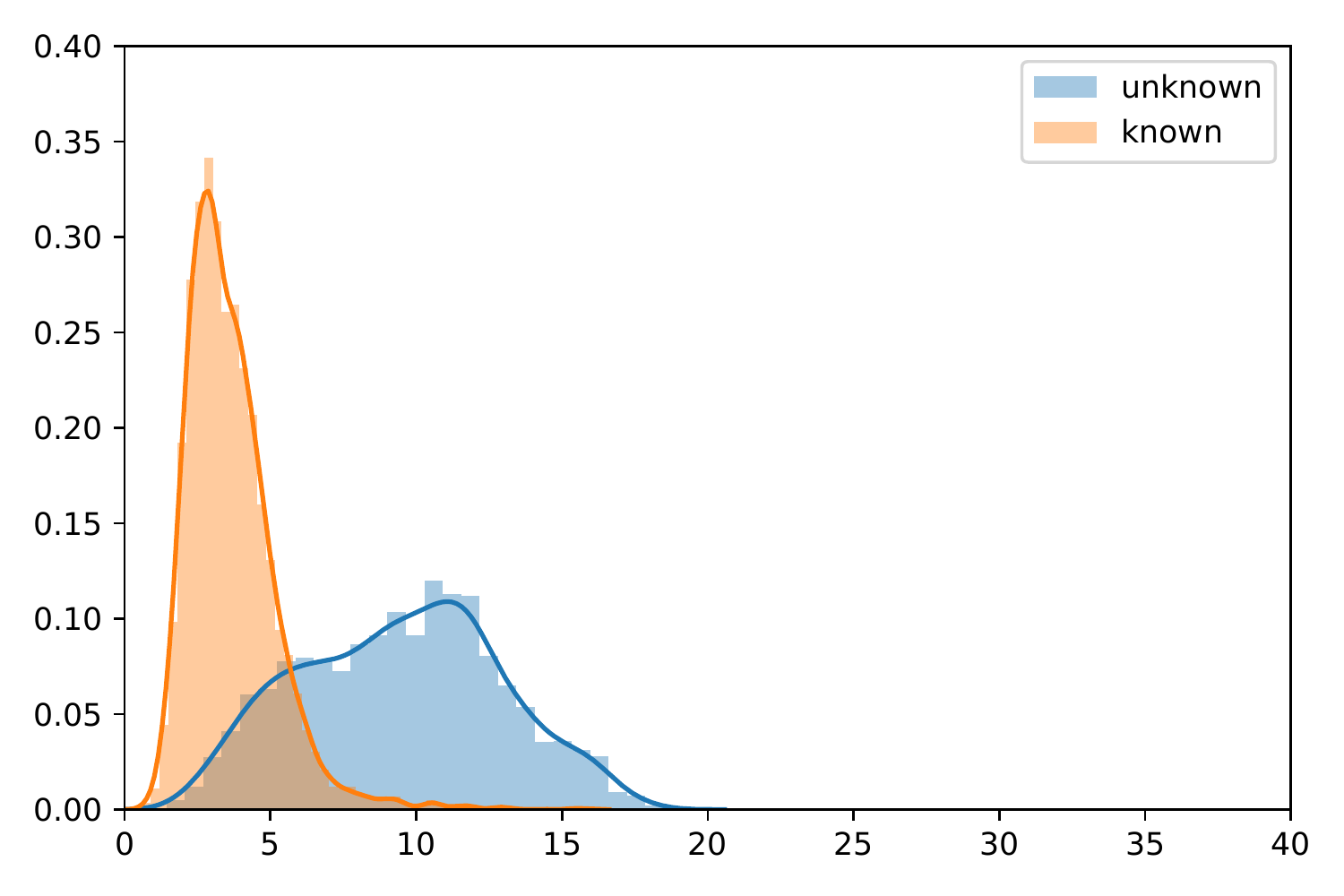}
                \caption{ce}
        \end{subfigure}%
  \begin{subfigure}[b]{0.25\textwidth}
                \includegraphics[width=\linewidth]{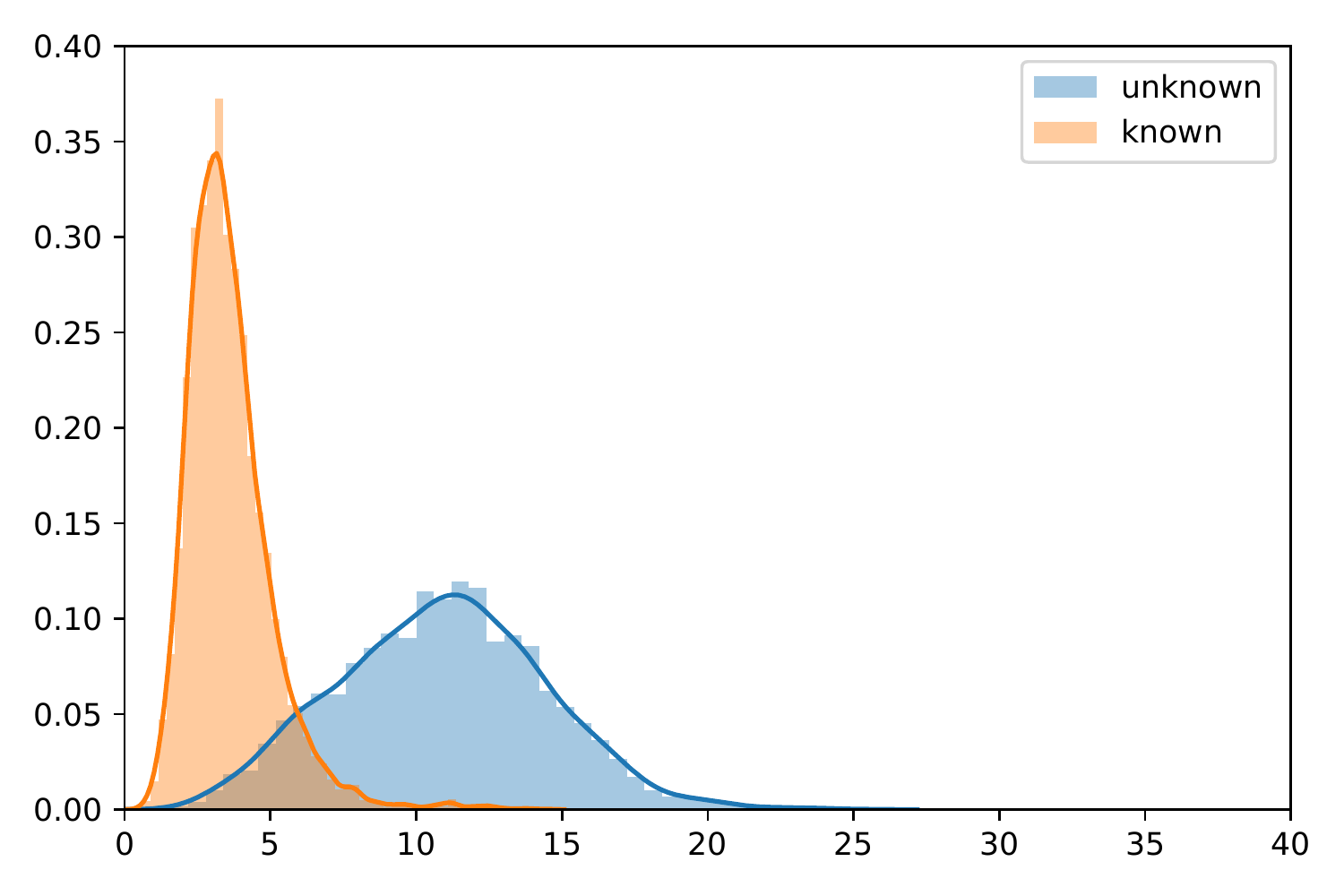}
                \caption{ce+MMF}
        \end{subfigure}%
 \\ \begin{subfigure}[b]{0.25\textwidth}
            \includegraphics[width=\linewidth]{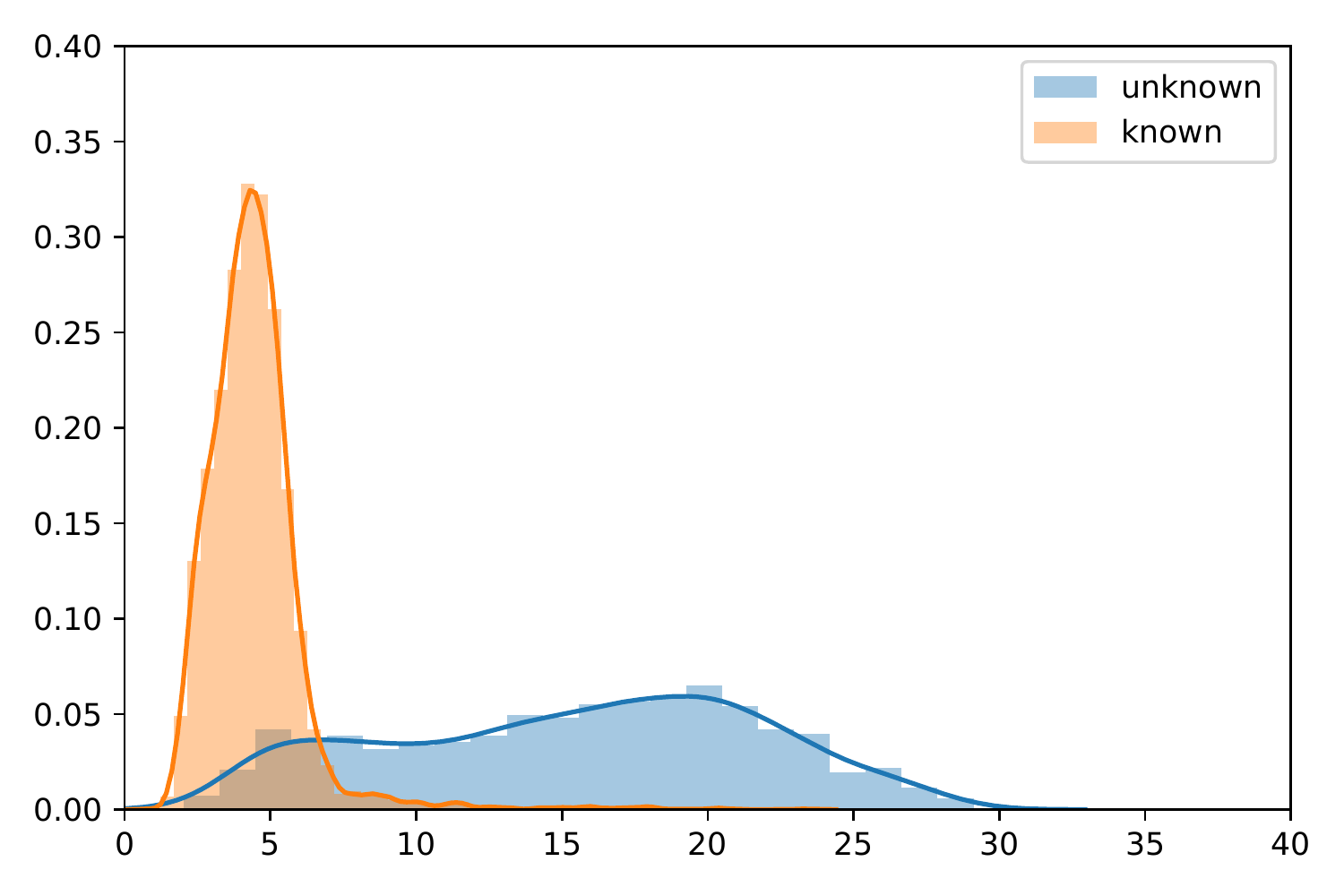}
                \caption{ii}
        \end{subfigure}%
  \begin{subfigure}[b]{0.25\textwidth}
                \includegraphics[width=\linewidth]{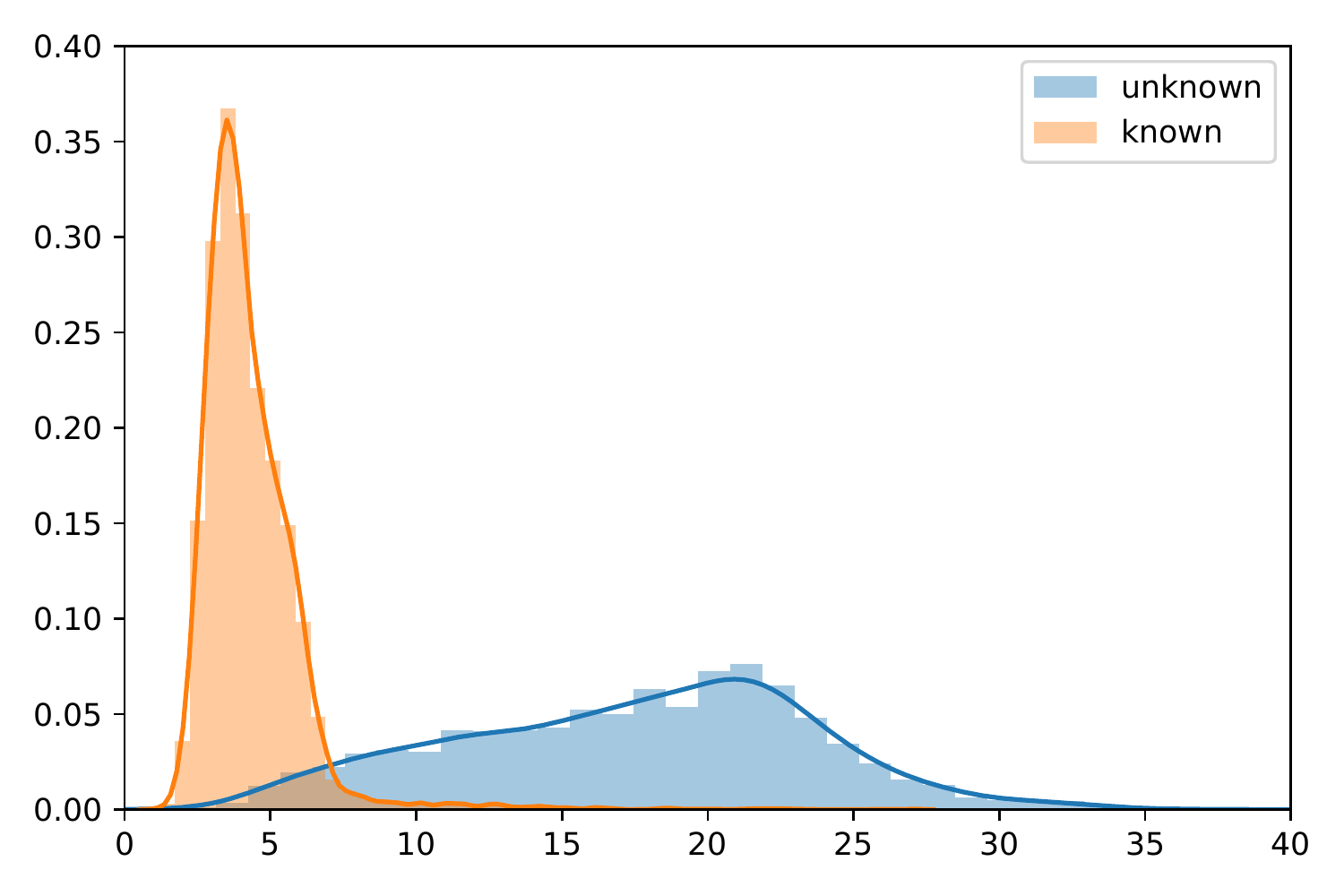}
\caption{ii+MMF}
        \end{subfigure}%
\caption{The distributions of outlier scores of the known and unknown
classes of the MNIST dataset in the experiments of ce vs.
ce+MMF and ii vs. ii+MMF.}
\label{fig: ce-hist}
\end{figure}

\begin{figure*}[]
 \begin{subfigure}[b]{0.5\textwidth}
               \includegraphics[width=\linewidth]{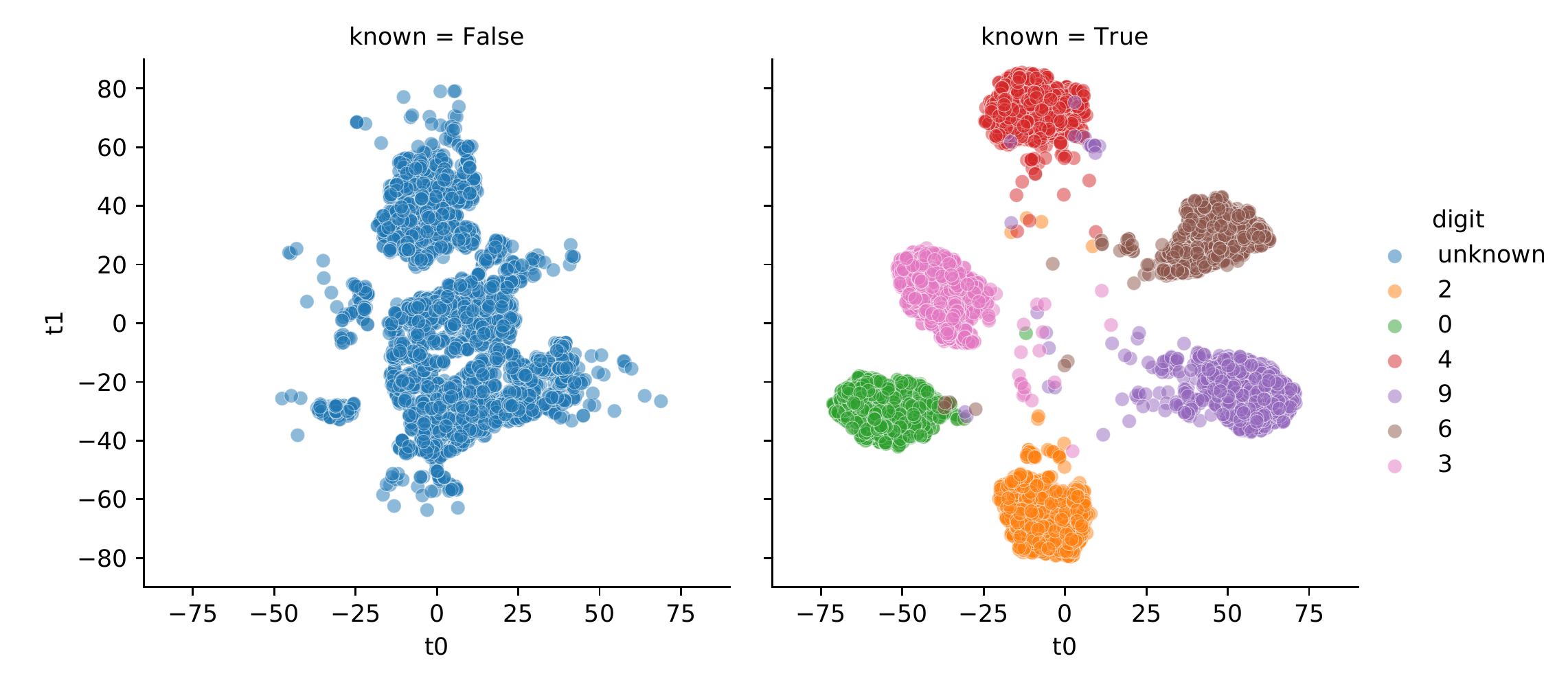}
                \caption{ce}
        \end{subfigure}%
  \begin{subfigure}[b]{0.5\textwidth}
  
                \includegraphics[width=\linewidth]{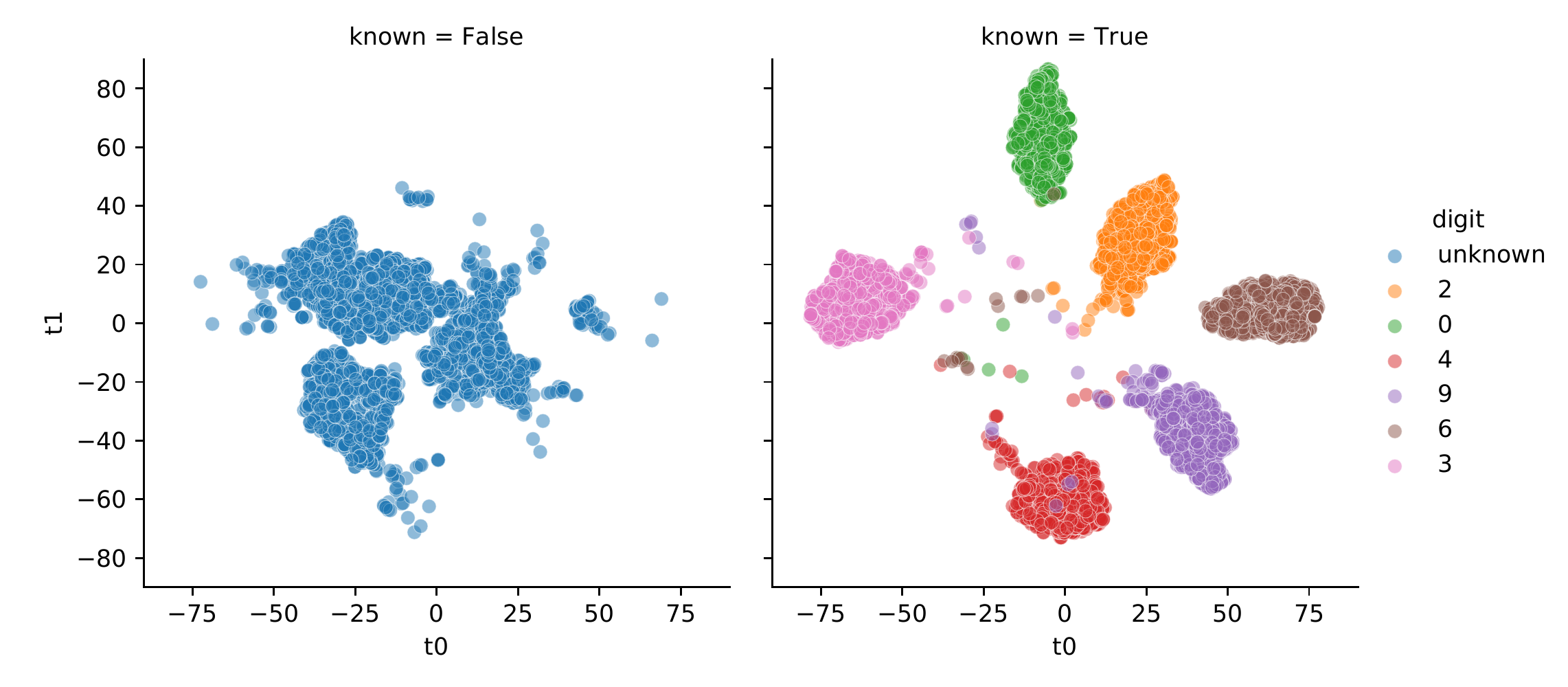}
                \caption{ce+MMF}
        \end{subfigure}%
 \\ \begin{subfigure}[b]{0.5\textwidth}
            \includegraphics[width=\linewidth]{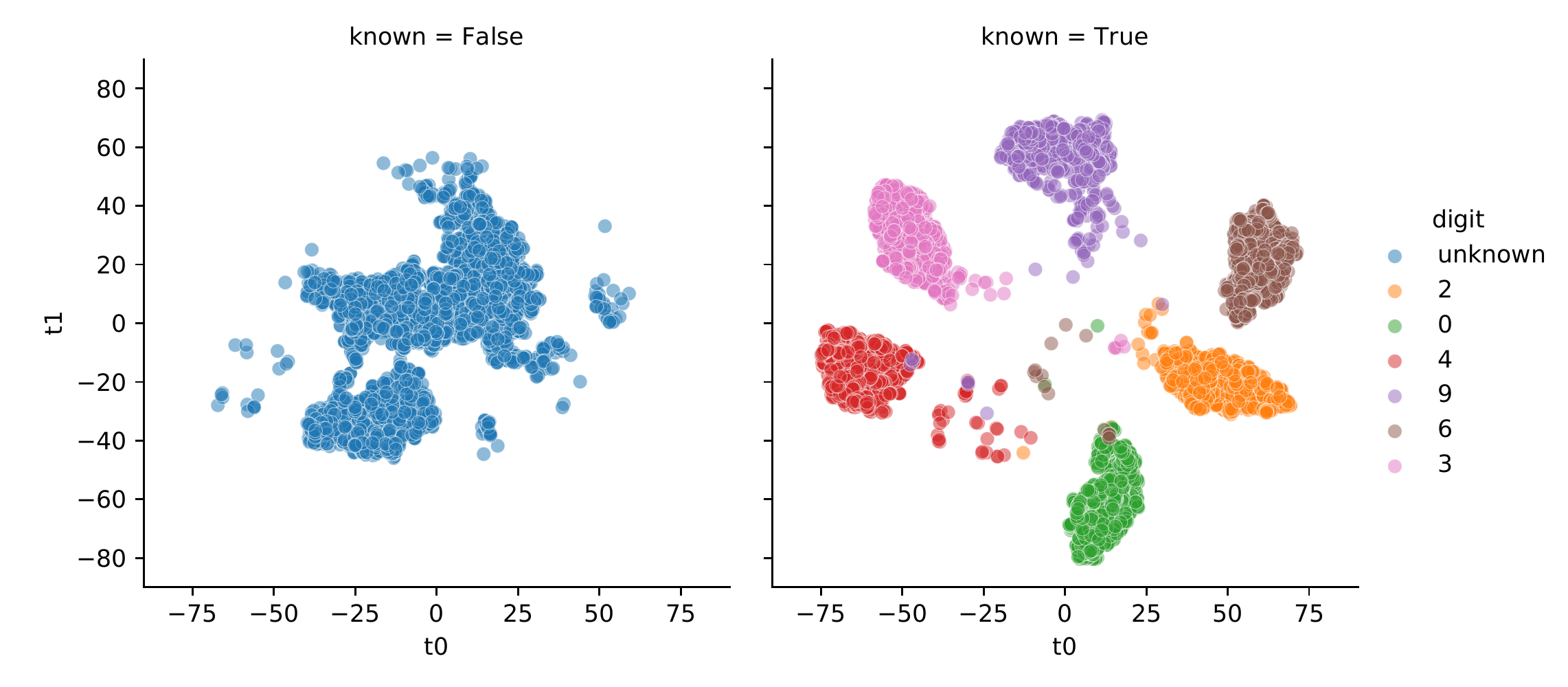}
                \caption{ii}
        \end{subfigure}%
  \begin{subfigure}[b]{0.5\textwidth}
                \includegraphics[width=\linewidth]{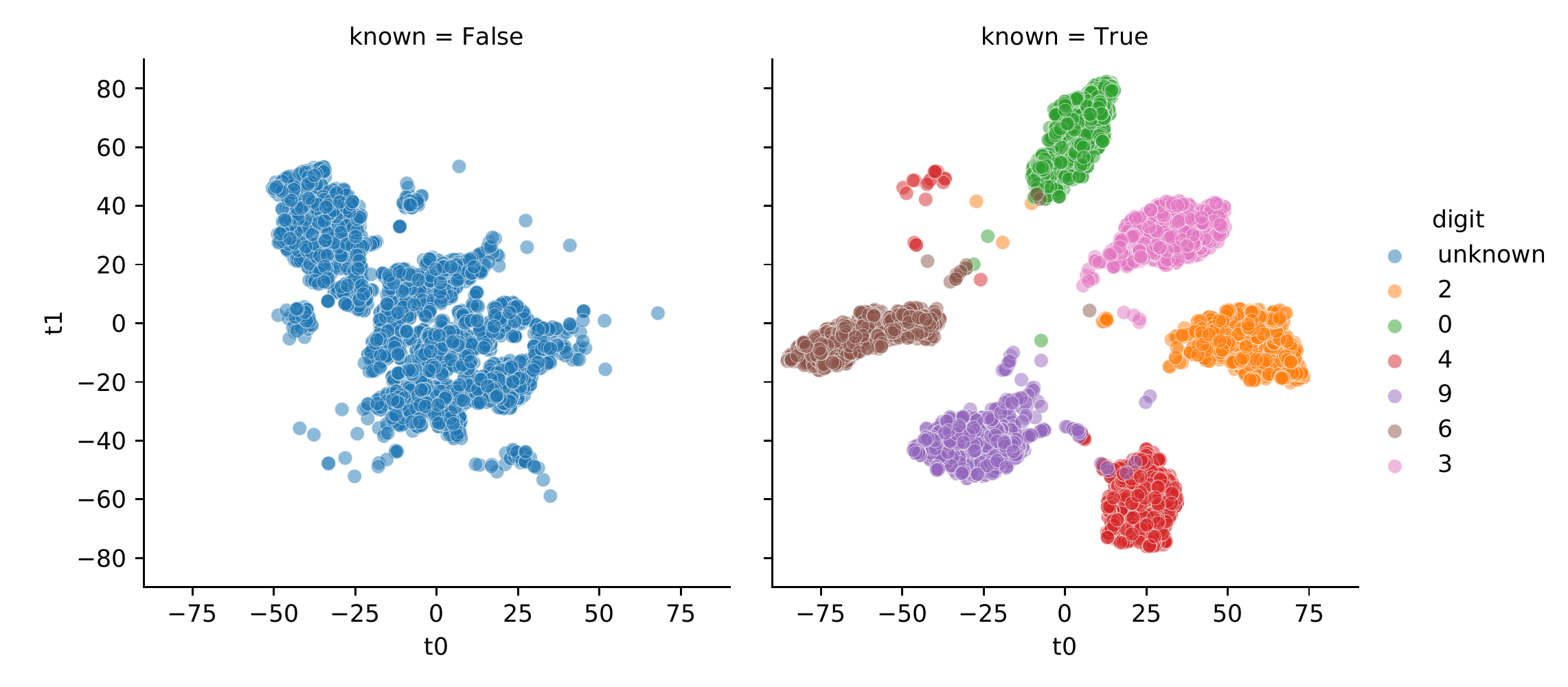}
\caption{ii+MMF}
        \end{subfigure}%
\caption{The t-SNE plots of the MNIST dataset in the experiments of ce vs. ce+MMF and ii vs. ii+MMF. The left subplots of (a) and (b) are the representations of the unknown class (a mixture of digits ``1'', ``5'', ``7'' and ``8''), and the right plots are the representations of the known classes.}
\label{fig: ce-tsne}
\end{figure*}

 In the paper, we analyzed the distributions of the outlier scores and the Z-layer representations of the MNIST dataset in the experiments of triplet loss and its association with the MMF extension. Figure \ref{fig: ce-hist} shows the histograms of the distributions of the outlier scores in the experiments of cross-entropy loss and ii loss and their associations with our MMF extension. We can see that same as its effects on the triplet loss in the paper, the MMF extension pushes the outlier scores of unknown samples further away from the known samples, reducing the overlaps between them. Figure \ref{fig: ce-tsne} shows the t-SNE (perplexity: 50) plots of the Z-layer representations of the MNIST dataset from the corresponding experiments. We observe that the representations of different classes become more separate after applying the MMF extension.

\begin{figure*}[]
\centering
\begin{subfigure}[b]{0.5\textwidth}
               \includegraphics[width=\linewidth]{Approach/mnist_cnn_ce_diff_heatmap1.pdf}
                \caption{ce}
                \label{heat-ce}
        \end{subfigure}%
  \begin{subfigure}[b]{0.5\textwidth}
                \includegraphics[width=\linewidth]{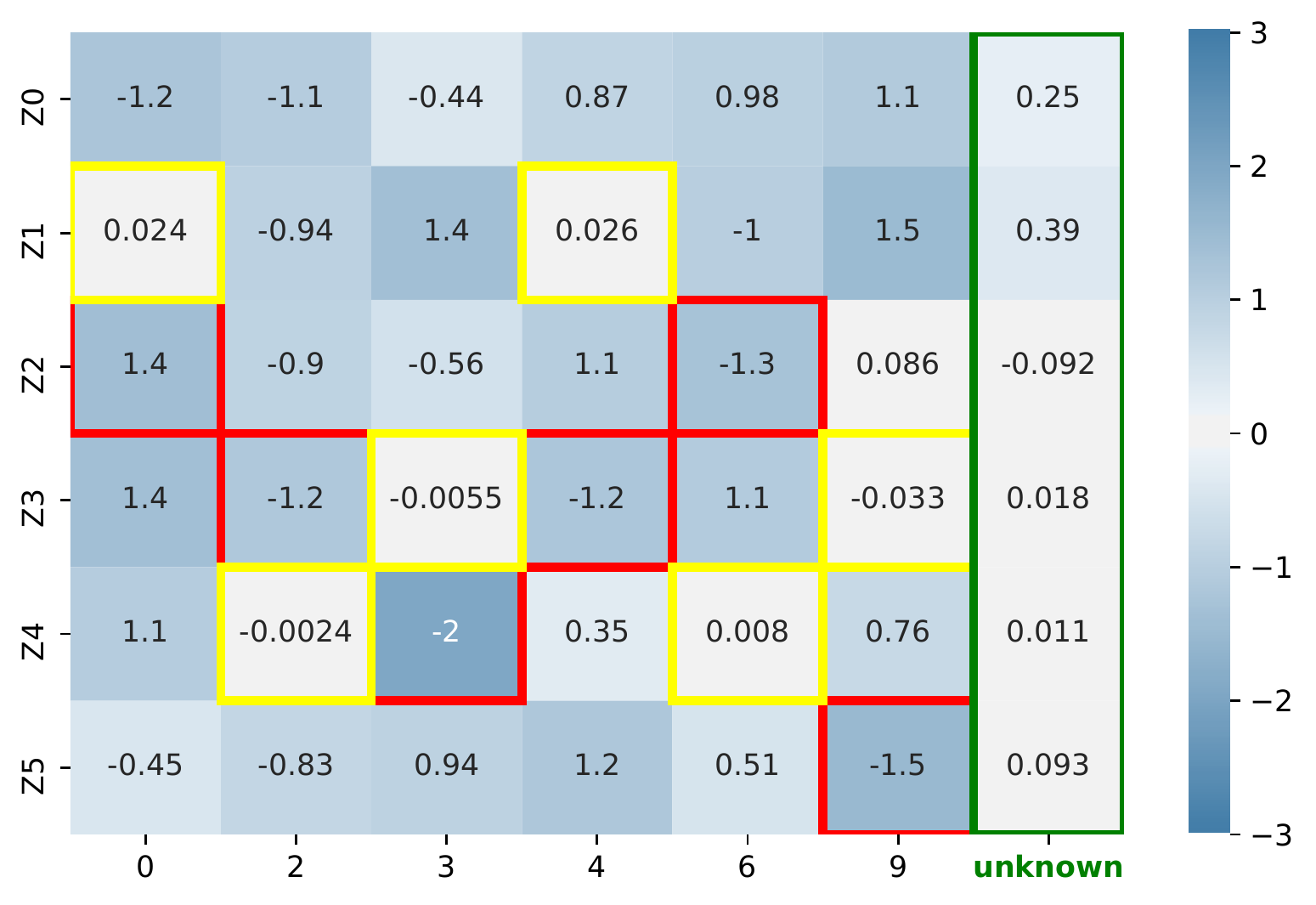}
                \caption{ce+MinF}
                \label{heat-ceminf}
        \end{subfigure}%
\\ \begin{subfigure}[b]{0.5\textwidth}
               \includegraphics[width=\linewidth]{Experiments/mnist_cnn_cemmf_max_diff_heatmap.pdf}
                \caption{ce+MaxF}
                \label{heat-cemaxf}
        \end{subfigure}%
   \begin{subfigure}[b]{0.5\textwidth}
                \includegraphics[width=\linewidth]{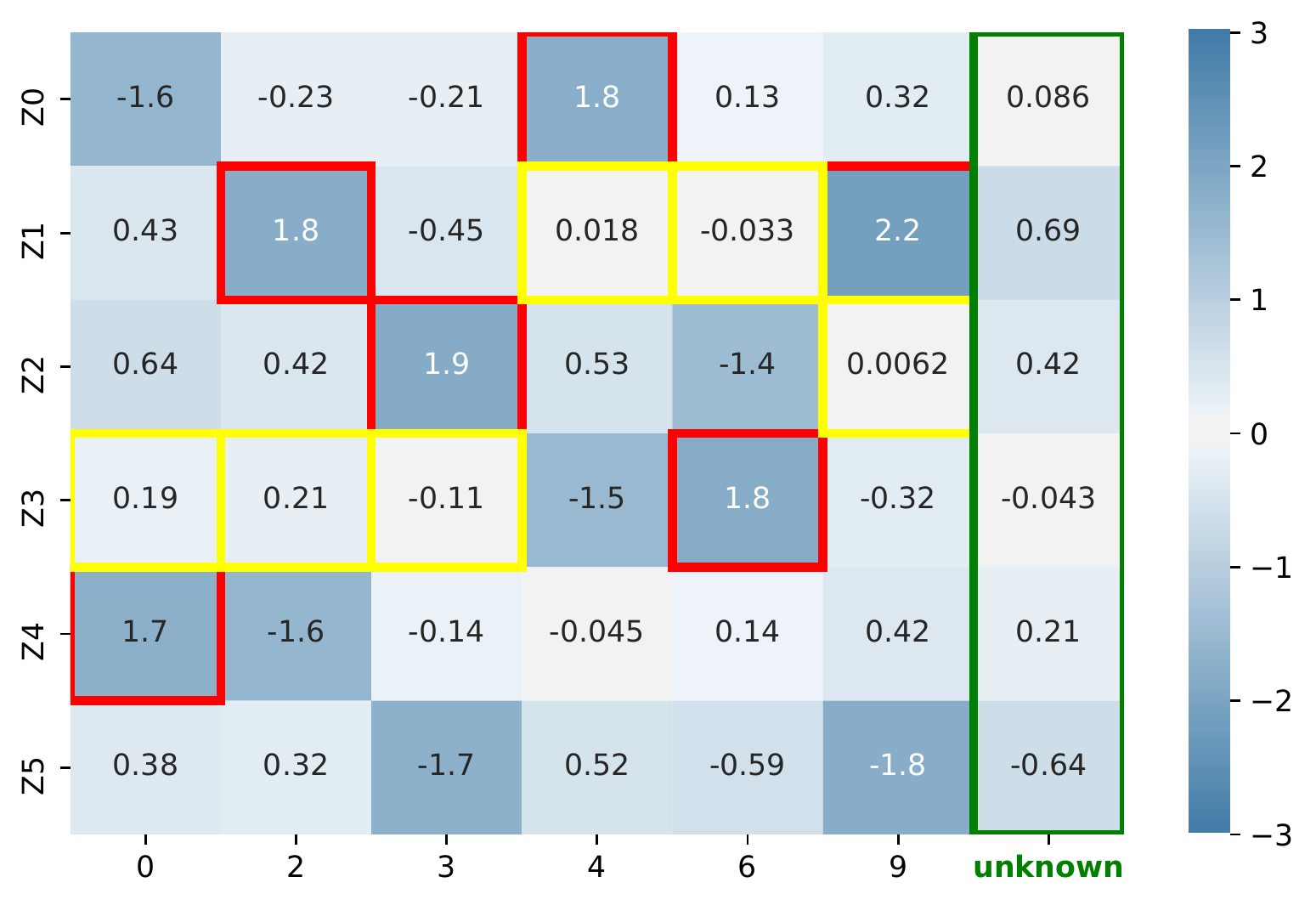}
                \caption{ce+MaxF2}
                \label{heat-cemaxf2}
        \end{subfigure}%
 \\\begin{subfigure}[b]{0.5\textwidth}
            \includegraphics[width=\linewidth]{Experiments/mnist_cnn_cemmf_diff_heatmap.pdf}
                \caption{ce+MMF}
                \label{heat-cemmf}
        \end{subfigure}%
  \begin{subfigure}[b]{0.5\textwidth}
                \includegraphics[width=\linewidth]{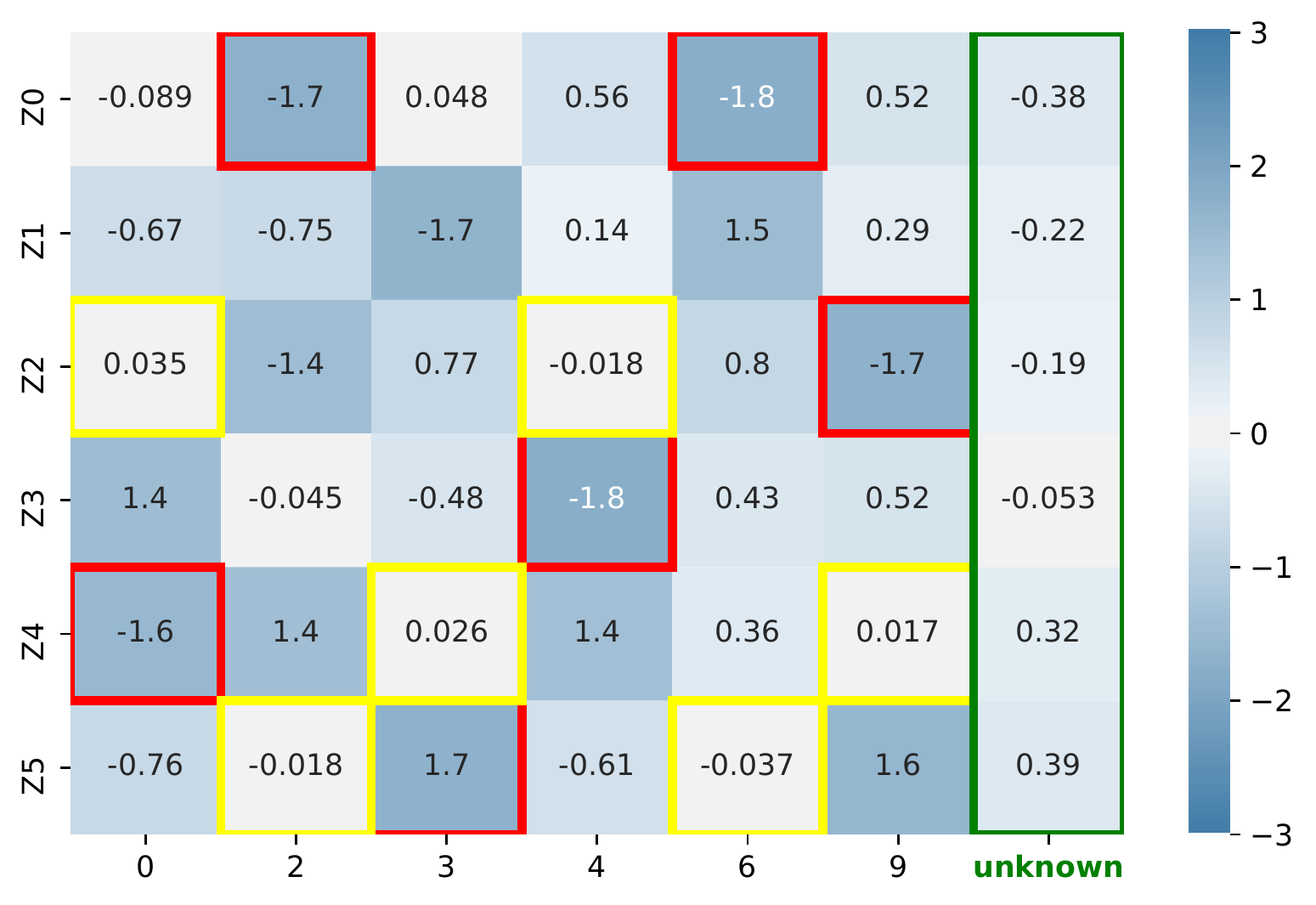}
\caption{ce+MMF2}
\label{heat-cemmf2}
        \end{subfigure}%
\caption{The heatmaps of MAV values of the MNIST dataset from different approaches.}
\label{fig: ce-heatmap}
\end{figure*}

 Figure \ref{fig: ce-heatmap} shows the heatmaps of the MAV values generated from different approaches on the MNIST dataset. The ones generated from cross-entropy only (ce) and the association with our MMF extension (ce+MMF) are already included in the paper. We can see that while the MaxF extension in Figure \ref{heat-cemaxf} emphasizes the most significant features, it finds a significant feature for each class. The MinF extension in \ref{heat-ceminf} helps maintain the least significant feature value for each class of a lower magnitude. Thus the MMF extension in Figure \ref{heat-cemmf} can find the significant features for each known class meanwhile keep the feature values of the unknown classes at a lower level to be better separated, which agrees with the results in Table 1 in the paper. 
 
 The heatmaps can help explain why ``dual max\_feature'' does not generally improve upon ``max\_feature''. Comparing Figures \ref{heat-cemaxf} and \ref{heat-cemaxf2}, we observe that the MaxF2 extension finds two significant features for each known class but of lower magnitudes, roughly half of those for the MaxF extension. The values in $max\_feature$ vector in MaxF are based on the most significant feature from each the known classes: (2.5, 2.6, -2.6, -2.7, -2.6, -2.5). Meanwhile the values in $max\_feature\_h$ in MaxF2 are from significant features: (1.7, 1.8, 1.9, 1.8, 1.8, 2.2), and those in $max\_feature\_l$ in MaxF2 are from significant features: (-1.6, -1.6, -1.7, -1.5, -1.4, -1.8). Similarly, comparing Figures \ref{heat-cemmf} and \ref{heat-cemmf2}, we observe that the magnitudes of significant features for the MMF2 extension are roughly half of those for the MMF extension. However, finding two significant features of lower magnitudes is not as helpful as finding one significant feature of a higher magnitude when we use Euclidean distance for computing the outlier scores. Since smaller magnitudes can yield lower outlier scores, instances from the unknown class can be more difficult to identify. Moreover, from Figure \ref{heat-cemaxf2} and Figure \ref{heat-cemmf2}, we observe that adding a MinF extension for ``dual max\_feature'' does not significantly lower the feature values of unknown classes.
 
\section{Datasets and Source Code}
As we mentioned in the paper, we use three datasets in two domains (image and malware) to evaluate our MMF extension. They are: MNIST \cite{lecun_cortes_burges}, Microsoft Challenge dataset (MC) \cite{DBLP:journals/corr/abs-1802-10135} and Android Genome (AG) \cite{zhou_jiang}. The MNIST dataset contains ten digits from 0 to 9, representing ten classes. The MC and AG datasets both contain nine malware families, representing nine classes. In the experiments, we split the datasets into training sets and test sets in the ratio 3:1. To simulate an open set scenario, we randomly select six classes as known classes and combine the rest four classes as an unknown class. The samples from the unknown class are excluded from the training set. The source code and preprocessed datasets are provided at Github: \href{https://github.com/clojia/mmf-openset}{https://github.com/clojia/mmf-openset}.

\section{Potential Future Work}
Our paper has been mainly focused on using a loss extension to make the learned features discriminative and representative. There are more ideas to be explored to solve OSR problems better. For example, we can use validation sets to find a more sophisticated threshold for the outlier scores. Explicitly, we can simulate an open-set dataset for the training set by leaving one class in the training set as unknown, thus if we have $C$ known classes, the training set should contain $C-1$ classes, and the validation set should contain $C$ classes. We can use a $C$-fold cross-validation to find the proper thresholds that maximize F1 scores. Finally, we can perform the training on all the $C$ known classes and use the average threshold as the final threshold.

Moreover, our current work identifies all the classes not covered by the training set as one unknown class. Future work concerns a further description of the unknown instances. i.e., the discovery of different novel classes. Once we detect novel classes, we can incorporate them into the model as known classes. Furthermore, recent work in open set recognition has shown promising results against adversarial attacks \cite{DBLP:conf/bmvc/RozsaGB17}, our work might also help identify adversarial instances and potentially be applied to adversarial attack detection systems.

\bibliographystyle{acm}
\bibliography{supplement}